\documentclass[lettersize,journal]{IEEEtran}
\usepackage{amsmath,amsfonts}
\usepackage{algorithmic}
\usepackage{algorithm}
\usepackage{array}
\usepackage[caption=false,font=normalsize,labelfont=sf,textfont=sf]{subfig}
\usepackage{textcomp}
\usepackage{stfloats}
\usepackage{url}
\usepackage{multicol}
\usepackage{booktabs}
\usepackage{multirow}
\usepackage[table]{xcolor}
\usepackage{verbatim}
\usepackage{graphicx}
\usepackage{hyperref}
\usepackage{cite}
\begin{document}

\title{SAMA: Semantic Anchor-aligned Augmentation for \\ Unified Low-Resource Multimodal Information Extraction}

\author{Quanjiang Guo, Chong Mu, Jiazhou Pan, Ming Jia, Ling Tian, Hui Gao, Zhao Kang,~\IEEEmembership{Member,~IEEE}
\thanks{This work was supported by the National Natural Science Foundation of China (No. U24A20323). Corresponding author: Zhao Kang.\\
\indent The source code and datasets are available at \href{https://github.com/UESTC-GQJ/SAMA}{https://github.com/UESTC-GQJ/SAMA}.\\
\indent The authors are with the School of Computer Science and Engineering, University of Electronic Science and Technology of China, Chengdu 611731, China (E-mail: guochance1999@163.com; \{muchong, jzhoupan\}@std.uestc.edu.cn; mjia7772@163.com; \{lingtian, huigao, zkang\}@uestc.edu.cn.)}
}

\markboth{IEEE TRANSACTIONS ON MULTIMEDIA,~Vol.~XX, No.~XX, XX~2025}%
{Shell \MakeLowercase{\textit{et al.}}: A Sample Article Using IEEEtran.cls for IEEE Journals}


\maketitle

\begin{abstract}
Multimodal Information Extraction (MIE)—covering tasks such as Multimodal Named Entity Recognition (MNER), Relation Extraction (MRE), and Event Extraction (MEE)—is essential for understanding multimedia content but remains constrained by severe data scarcity. Although data augmentation is a promising remedy, existing approaches are impeded by coarse cross-modal alignment and fragmented, task-specific designs that fail to exploit shared semantic knowledge. To overcome these limitations, we introduce Semantic Anchor-aligned Multimodal Augmentation (SAMA), a unified framework for generating high-fidelity, task-aware synthetic data. SAMA constructs structured semantic anchors from ground-truth labels to guide a Collaborative Multi-Experts Multimodal Large Language Model (CME-MLLM), which integrates a Universal Adapter for shared semantics with Task-Specific Adapters to produce diverse yet constraint-compliant textual samples. For image synthesis, SAMA employs an Anchor-Preserving Diffusion mechanism that uses anchor-weighted prompts and latent conditioning to maintain critical semantic anchors while diversifying visual contexts. To eliminate the need for manual verification, SAMA further introduces a Dual-Constraint Filtering module that selects synthetic samples based on both cross-modal consistency and anchor fidelity. Extensive experiments across benchmark datasets for MNER, MRE, and MEE demonstrate that SAMA consistently outperforms state-of-the-art augmentation baselines under both fully supervised and low-resource settings, underscoring its versatility, robustness, and effectiveness.
\end{abstract}

\begin{IEEEkeywords}
Multimodal Large Language Models, Synthetic Data Generation, Cross-Modal Alignment, Diffusion Models.
\end{IEEEkeywords}


\section{Introduction}

\begin{figure}[!htbp]
  \centering
    \includegraphics[width=1\linewidth]{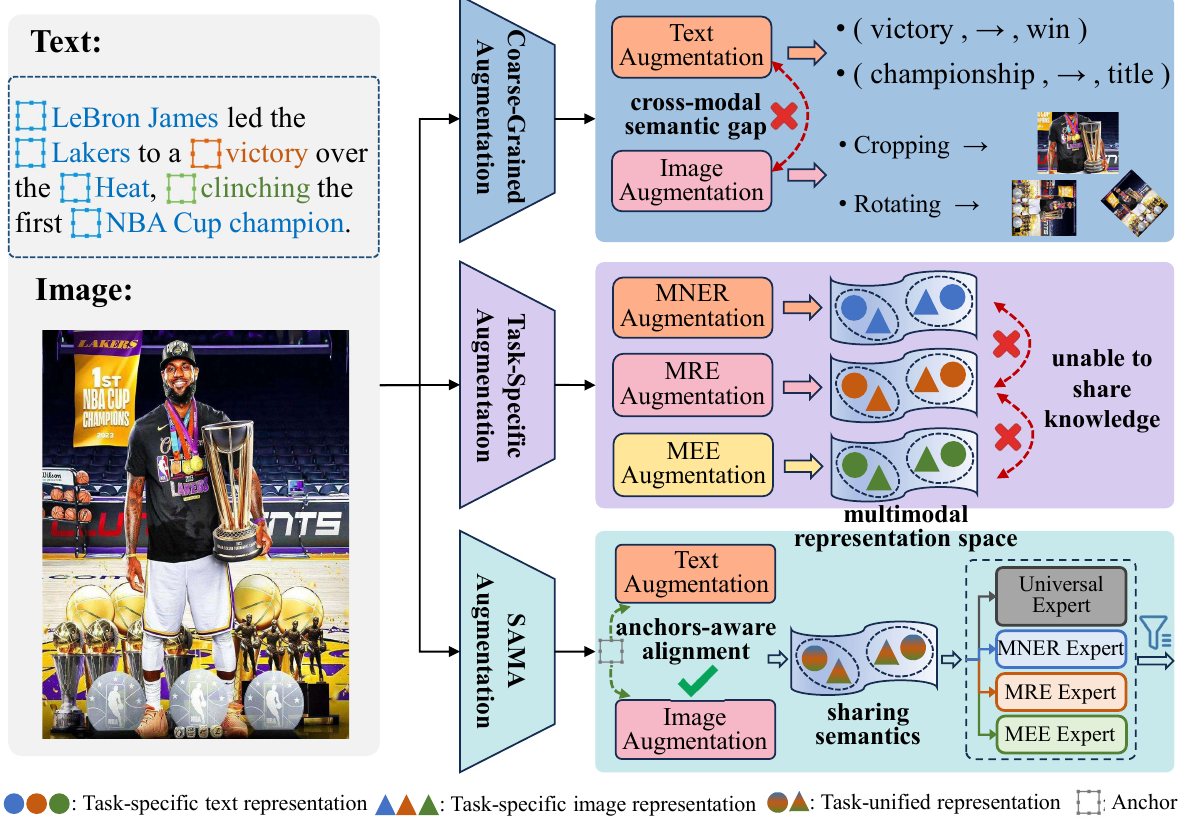}
  \caption{The comparison of existing data augmentation methods and our SAMA. (a) \textbf{Coarse-grained DA methods}: Independent text-image augmentation with cross-modal semantic gaps. (b) \textbf{Task-specific DA methods}: Reliance on task-specific architectures that hinder knowledge sharing. (c) \textbf{Our SAMA framework}: Unified multi-expert generation with semantic anchor-aware cross-modal alignment.}
  \label{fig:motivation}
\end{figure}

\IEEEPARstart{I}{nformation} extraction (IE), a task focused on extracting structured information from unstructured texts, is a fundamental component of natural language processing. The increasing prevalence of multimodal content on social platforms such as Twitter and Instagram has spurred the need for Multimodal Information Extraction (MIE), which aims to identify structured knowledge (e.g., entities, relations, events, and attributes) from text-image pairs and facilitate the construction of multimodal knowledge graphs~\cite{lu2018visual, moon2018multimodal,zheng2021multimodal,jia2024mujo,zhu2024fcds,guo2025bridging}. 

\begin{figure}[!htbp]
  \centering
    \includegraphics[width=0.8\linewidth]{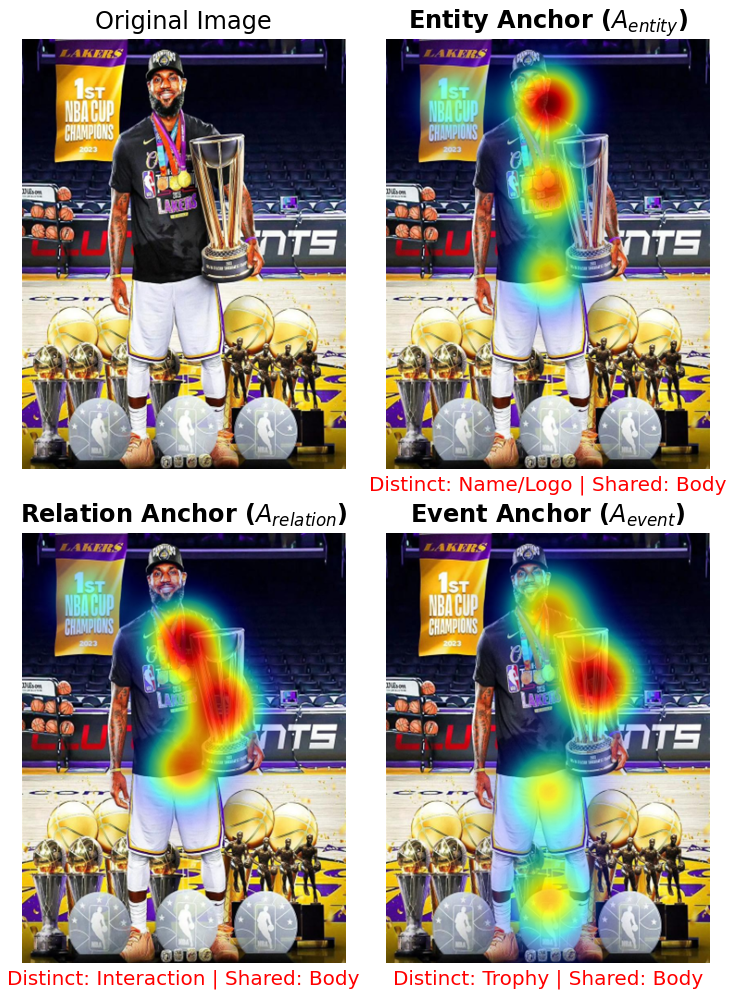}
  \caption{\textbf{Visualization of attention maps guided by different task anchors.} While distinct anchors (e.g., Entity, Relation) direct the model's focus to specific semantic regions (Distinct Focus), a consistent high-attention region persists on the main subject across all tasks (Shared Focus). This observed semantic overlap validates the feasibility of cross-task knowledge sharing and justifies the design of the Universal Adapter in SAMA.}
  \label{fig:motivation1}
\end{figure}

\sloppy While tasks such as Multimodal Named Entity Recognition (MNER), Relation Extraction (MRE), and Event Extraction (MEE) have seen advances in model architectures~\cite{li2023prompting,wang2022named,li2020cross,li2022clip}, their effectiveness in low-resource scenarios remains severely limited by two critical challenges. 1) \textbf{Data Scarcity and Cross-Modal Alignment}: Current MIE methods rely heavily on annotated multimodal datasets, where fine-grained labels are costly to acquire~\cite{zhang2018adaptive,li2024generative}. As illustrated in Figure~\ref{fig:motivation}, traditional Data Augmentation (DA) techniques, such as textual synonym replacement~\cite{csahin2018data} or image transformations, fail to generate semantically consistent text-image pairs, as they operate in isolation per modality. This exacerbates the cross-modal semantic gap, where textual entities and visual regions lack explicit alignment~\cite{dai2020analysis, zheng2020object}. 2) \textbf{Task-Specific Fragmentation and Limited Generalization}: Existing approaches often adopt task-specific architectures~\cite{zheng2021multimodal,zhao2022learning,liu2023document,guo-etal-2025-baner,guo2025extracting}, which hinder knowledge sharing across tasks. For instance, models trained for MNER cannot leverage event schemas from MEE, despite shared semantic structures. As shown in Figure~\ref{fig:motivation1}, our preliminary analysis reveals a crucial insight: while different task anchors guide the model to focus on task-specific regions, we observe a consistent high-attention region on the subject's body across all tasks. This validates that MNER, MRE, and MEE share a common foundation of semantic understanding, which suggests that a unified framework could effectively leverage these shared representations.

To address these challenges, we propose \textbf{S}emantic \textbf{A}nchor-aligned \textbf{M}ultimodal \textbf{A}ugmentation (\textbf{SAMA}), a novel framework that generates high-quality, task-agnostic multimodal data through \textit{collaborative multi-expert text generation} and \textit{anchor-perserving image synthesis}. In contrast to prior approaches that employ coarse-grained multimodal fusion~\cite{sun2024umie} or task-specific DA methods~\cite{li2024generative}, our approach uniquely bridges the gap between structured semantic units and cross-modal consistency, enabling robust augmentation for MNER, MRE, and MEE tasks under extreme data scarcity.

\begin{figure*}[t]
\centering
\includegraphics[width=0.98\linewidth]{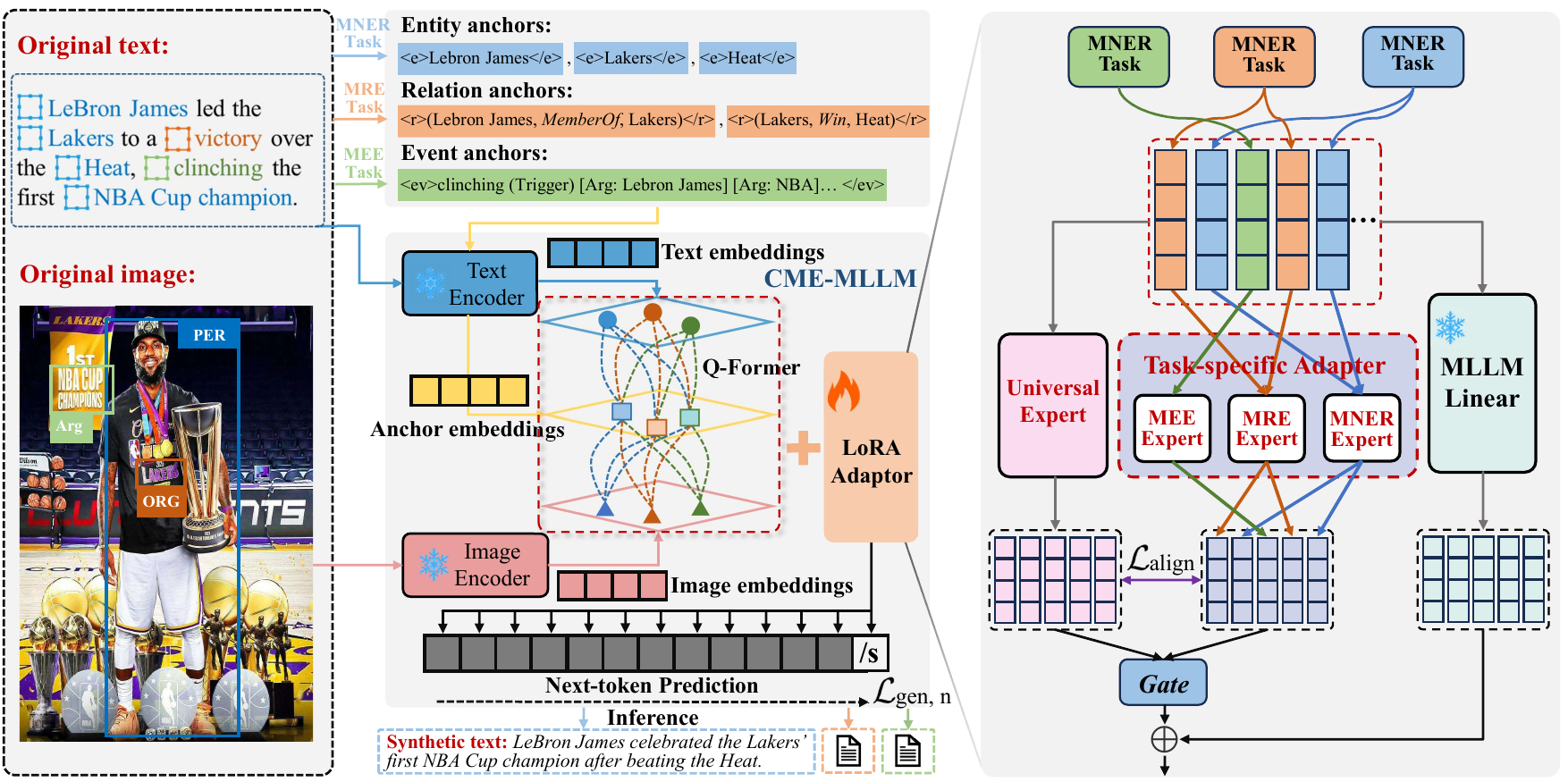}
\caption{Illustration of the semantic anchor-driven text generation pipeline in SAMA. Given an input text-image pair, we first extract structured \textbf{Semantic Anchors} that capture the essential semantic core. These anchors, combined with the visual context, condition the \textbf{Collaborative Multi-Experts MLLM}. Within this module, a shared \textbf{Universal Adapter} captures general cross-task knowledge, while distinct \textbf{Task-Specific Adapters} handle the fine-grained requirements for MNER, MRE, and MEE. A dynamic gating mechanism integrates outputs from these diverse experts to generate \textbf{Synthetic Text} that is contextually rich while strictly preserving the predefined anchor semantics.}
\label{Fig:overview}
\end{figure*}

Specifically, given a low-resource training sample, we first construct semantic anchors by re-encoding its ground-truth labels into a structured textual format. These anchors condition a Collaborative Multi-Experts MLLM (CME-MLLM), adapted from InstructBLIP~\cite{dai2023instructblip}. To exploit the shared semantics, this model synergizes a Universal Adapter for shared knowledge with Task-Specific Adapters to generate diverse synthetic text while strictly adhering to task constraints. For image synthesis, we employ an Anchor-Preserving Diffusion strategy. Unlike standard generation, this process is conditioned on both anchor-weighted prompts and the original image latents, ensuring that the visual identity of key entities is preserved while the background context is diversified. Finally, we employ a label projection strategy to automatically assign annotations to the synthetic pair, followed by a \textit{Dual-Constraint Filtering mechanism} that selects high-quality samples based on rigorous cross-modal alignment and anchor fidelity checks.

In summary, our contributions are as follows:

\begin{itemize}
    \item We propose \textbf{SAMA}, a unified generative framework for MNER, MRE, and MEE. By incorporating a \textbf{Collaborative Multi-Experts} architecture, it effectively harmonizes shared universal knowledge with task-specific features to achieve robust, task-agnostic generalization.
    \item We introduce a \textbf{semantic anchor-driven} paradigm that conditions both the adapter-based text generation and \textbf{anchor-preserving image synthesis}. This mechanism strictly enforces cross-modal consistency and prevents the semantic drift often seen in generative augmentation.
    \item We design a \textbf{Dual-Constraint Filtering} mechanism to guarantee the reliability of synthetic data. Extensive experiments demonstrate that SAMA significantly outperforms state-of-the-art baselines in both low-resource and fully supervised settings.
\end{itemize}

\section{Related Work}
\subsection{Multimodal Named Entity Recognition}
Multimodal Named Entity Recognition (MNER) aims to recognize named entities in text and classify them into predefined categories based on text-image pairs. Pioneering works~\cite{lu2018visual,moon2018multimodal,zhang2018adaptive} focus on fusing visual information for improved word representation learning. With the use of the multimodal transformer architecture, a variety of attention-based mechanisms~\cite{chen2023learning,chen2022hybrid,wu2023mcg,wu2020multimodal,xu2022maf,yu2020improving,li2025adaptive} are designed to model the interactions between textual and visual modalities. In addition, converting images into natural languages~\cite{wang2022ita} and retrieving external knowledge~\cite{li2023prompting,wang2022named} are used to enhance the textual information. Different from sequence labeling-based methods mentioned above, machine reading comprehension (MRC)-based methods~\cite{bao2023mpmrc,jia2022query}, in-context learning-based methods~\cite{cai2023context,chen2023chain}, index generation-based methods~\cite{yu2023grounded}, and paraphrase generation-based methods~\cite{wang2023fine} have been recently adapted to the MNER task.

\subsection{Multimodal Relation Extraction}
Multimodal Relation Extraction (MRE)~\cite{zheng2021mnre} aims to identify the semantic relationships between two entities based on the given text image pair. HVPNeT~\cite{chen2022good} fuses image information as a prefix for better text representation. MoRe~\cite{wang2022named} retrieves related texts from the entire Wikipedia dumps for boosting both MNER and MRE performance. Follow-up work~\cite{hu2023multimodal} further retrieves relevant images to the object, text, and image for better retrieval augmentation~\cite{xie2023adaptive,yue2023automatic}. However, they involve time-consuming retrieval over a large-scale collection and require an external knowledge base.

\subsection{Multimodal Event Extraction}
Multimodal Event Extraction (MEE) aims to extract events (i.e., Event Detection) and arguments for the event (i.e., Event Argument Extraction) from multiple modalities. Li et al.~\cite{li2020cross} propose the $\mathrm{M^2E^2}$ dataset and the WASE method, which uses weakly supervised learning to encode structured representations from textual and visual data into a shared embedding space. To better bridge the modalities, Unicl~\cite{liu2022multimedia} introduces a contrastive learning framework. More recently, MGIM~\cite{liu2024multi} proposes a multi-grained gradual inference mechanism, achieving fine-grained text-image alignment through iterative reasoning rounds.

However, all these methods above heavily rely on a large amount of annotated data, which requires fine-grained annotation of entity/relation/event for MIE. Since obtaining such human annotation is time-consuming and costly, our work aims to propose an effective data augmentation method for MIE to enrich the annotated text-image pairs, particularly in low-resource settings.

\subsection{Data Augmentation}
Data Augmentation (DA) is a pivotal strategy for mitigating data scarcity, evolving from early rule-based techniques like word replacement and span rotation~\cite{wei-zou-2019-eda, dai2020analysis} to generative approaches that better preserve semantic coherence~\cite{ding2020daga}. In the multimodal domain, methods have progressed from coarse-grained strategies like MixGen~\cite{hao2023mixgen}, which interpolates images and text, to sophisticated generative frameworks tailored for specific extraction tasks. For instance, GMDA~\cite{li2024generative} utilizes a two-stage pipeline to synthesize aligned text-image pairs specifically for MNER, while CAMIM~\cite{zhang2024caption} leverages caption-aware cross-attention to enhance fine-grained alignment for MNRE. Similarly, CAMEL~\cite{du2023training} focuses on augmenting MEE by generating missing modalities through iterative training. However, these approaches often rely on task-specific architectures or separate generation pipelines, limiting the potential for sharing common semantic knowledge across different MIE subtasks. Moreover, while closed-source commercial Multimodal LLMs demonstrate impressive general capabilities, they are suboptimal for low-resource MIE augmentation due to: 1) \textbf{Prohibitive Costs and Latency}: Augmenting entire datasets incurs unsustainable API costs and rate limits. 2) \textbf{Lack of Schema Adherence}: Commercial models often prioritize descriptive fluency over the strict, fine-grained structural constraints (e.g., specific relation triplets) required by MIE tasks. SAMA provides a cost-effective, open-source alternative that aligns latent representations directly with specific task schemas. To address this, our work proposes a unified, anchor-aligned framework capable of generating high-fidelity data for MNER, MRE, and MEE simultaneously.

\section{Methodology}

In this section, we introduce the \textbf{Semantic Anchor-aligned Multimodal Augmentation (SAMA)} framework, designed to generate high-quality text-image pairs and uses these to augment the training data, improving MIE in low-resource settings.

\subsection{Problem Formulation}

MIE involves predicting structured information from paired text and image data. Formally, given a multimodal input consisting of a text sequence $\mathbf{T} = \{w_1, \dots, w_L\}$ and an image $\mathbf{I}$, the objective is to predict a structured output set $\mathbf{Y} = \mathcal{E} \cup \mathcal{R} \cup \mathcal{V}$, where $\mathcal{E}$, $\mathcal{R}$, and $\mathcal{V}$ denote the sets of entities, relations, and events, respectively.

In this work, we target low-resource scenarios, where only a limited labeled dataset $\mathcal{D}_{\text{train}} = \{(\mathbf{T}_i, \mathbf{I}_i, \mathbf{Y}_i)\}_{i=1}^N$ is available. The goal of SAMA is to synthesize a high-fidelity augmented dataset $\mathcal{D}_{\text{syn}} = \{(\hat{\mathbf{T}}_j, \hat{\mathbf{I}}_j, \hat{\mathbf{Y}}_j)\}_{j=1}^K$ that preserves the semantic consistency of the original labels $\mathbf{Y}$ while introducing necessary linguistic and visual diversity to improve model generalization.

\subsection{Preliminaries}
SAMA is built upon InstructBLIP, a Multimodal Large Language Model (MLLM). InstructBLIP employs a Querying Transformer (Q-Former) to align visual features from a frozen image encoder with a frozen Large Language Model (LLM).

To adapt this architecture for task-specific generation without catastrophic forgetting, we utilize Low-Rank Adaptation (LoRA)~\cite{hulora}. LoRA freezes the pre-trained model weights $W_0$ and injects trainable rank decomposition matrices $A$ and $B$ into the linear layers:
\begin{equation}
h = W_0 x + B A x ,
\end{equation}
where $A \in \mathbb{R}^{r \times d_{in}}$, $B \in \mathbb{R}^{d_{out} \times r}$, and $r \ll d_{in}$. In SAMA, we employ LoRA not just for efficient fine-tuning, but as the structural basis for our Collaborative Multi-Experts mechanism, where different sets of LoRA matrices ($A, B$) represent the Universal and Task-specific adapters.

\subsection{Semantic Anchor Construction}
Semantic anchors ($\mathcal{A}$) serve as the unified control signal for the SAMA framework. Rather than directly extracting information from the input text, we construct these anchors by re-encoding the ground-truth labels $\mathbf{Y}_i$ from the low-resource training set into a structured textual format. 
From an information-theoretic perspective, this inline structured tagging acts as a hard constraint that minimizes the conditional entropy $H(\mathbf{Y}|\mathbf{X})$ of the generated sequence. Unlike external instructions (e.g., JSON schemas), which primarily impose soft constraints, inline tags dynamically restrict the probability mass of the next-token distribution to the valid schema space during autoregressive decoding.

\subsubsection{Structured Textual Tagging}
As illustrated in Figure~\ref{Fig:overview}, we parse the annotations into task-specific anchor strings using explicit markup tokens to enforce semantic constraints:

\begin{itemize}
\item \textbf{Entity Anchors (MNER)}: We wrap entity spans with type-specific tags (e.g., $\langle \texttt{PER} \rangle \text{LeBron James} \langle /\texttt{PER} \rangle$) to explicitly bind the text to its entity type, preventing type drift.
\item \textbf{Relation Anchors (MRE)}: Relations are linearized into strict subject-relation-object triples (e.g., $\langle \text{Subject}, \texttt{MemberOf}, \text{Object} \rangle$). This structure captures the directionality of interactions.
\item \textbf{Event Anchors (MEE)}: We employ a hierarchical format (e.g., $\langle \texttt{Event:Win} \rangle \text{Trigger} \langle \texttt{Arg:Winner} \rangle \text{ArgName}$) that links event triggers with their arguments, preserving the dependency structure between the action and its participants.
\end{itemize}




\subsubsection{Unified Multimodal Encoding}
To utilize these discrete textual anchors as continuous control signals, we encode them alongside the visual context. For a specific task $n \in \{\text{MNER}, \text{MRE}, \text{MEE}\}$, the task-specific anchor set $\mathcal{A}_n$ is processed through the InstructBLIP fusion layer:
\begin{equation}
\mathbf{s}_{\text{anchor}, n} = f_{\text{InstructBLIP}}(\mathbf{T}, \mathbf{I}, \mathcal{A}_n) \in \mathbb{R}^d
\end{equation}

This yields a task-aware anchor embedding $\mathbf{s}_{\text{anchor}, n}$ that fuses textual semantics with visual features, serving as the condition for the collaborative multi-experts mechanism.

\subsection{Collaborative Anchor-guided Generation}
SAMA leverages an \textbf{CME-MLLM} to generate diverse, yet consistent text. This model is empowered by a collaborative multi-experts mechanism that dynamically balances shared multimodal knowledge with task-specific constraints.

\subsubsection{Collaborative Multi-Experts Adapters}
To effectively leverage the semantic anchors for generation, we introduce a collaborative adapter mechanism within the \textbf{CME-MLLM)}. For a given input sample belonging to task $n \in \{\text{MNER}, \text{MRE}, \text{MEE}\}$, two distinct LoRA-based experts are activated:

\begin{itemize}
\item \textbf{Universal Adapter} $U(x)$: This component encodes shared multimodal knowledge. It is optimized jointly across all tasks to capture common semantic regularities and shared representational structures (e.g., the consistent allocation of attention to primary visual entities across MNER, MRE, and MEE, as illustrated in Figure~\ref{fig:motivation1}).
\item \textbf{Task-Specific Adapter} $D_n(x)$: This component encodes features specialized for task $n$. It is trained exclusively on data from its associated task domain to model fine-grained, task-specific requirements (e.g., relational interactions in MRE or particular trigger–argument dependencies in MEE).
\end{itemize}


These adapters are integrated via an anchor-motivated gate mechanism. Let $x_t = [\mathbf{h}_{\text{text}}^{(t)}; \mathbf{h}_{\text{image}}^{(t)}]$ represent the concatenated features for the current context. The gate dynamically computes mixing weights $g = [g_u, g_d]$ based on both $x_t$ and the task-aware anchor embedding $\mathbf{s}_{\text{anchor}, n}$:
\begin{equation}
g = \mathrm{Softmax}\Big(W_g \cdot \big[ \mathbf{s}_{\text{anchor}, n} , x_t \big]\Big) ,
\end{equation}

where $W_g$ is a trainable projection matrix. The final hidden state $h_t$ is computed as the weighted sum of the expert outputs:
\begin{equation}
h_t = g_u \cdot U(x_t) + g_d \cdot D_n(x_t)
\end{equation}

Finally, this anchor-gated representation is projected into the vocabulary space to predict the next token probability:
\begin{equation}
P(\hat{w}_t \mid \hat{w}{<t}, \mathbf{T}, \mathbf{I}, \mathcal{A}_n) = \mathrm{Softmax}\Big(\mathbf{W}_{\text{head}} \cdot h_t\Big)
\end{equation}

where $\mathbf{W}_{\text{head}}$ denotes the weights of the LLM head.

\subsubsection{Knowledge Harmonization via Mutual Information}
To prevent the task-specific adapters $D_n(x)$ from overfitting to limited data, we employ a teacher-student knowledge alignment strategy inspired by~\cite{YuanCSLHD025}. We treat the Universal Adapter $U(x)$ as a ``teacher" that imparts essential shared semantics to the ``student" Task-Specific Adapters $D_n(x)$. We maximize the mutual information between their representations to facilitate knowledge transfer. The alignment loss is defined as:
\begin{equation}
\mathcal{L}_{\text{align}} = -\sum_{n} \sum_{i} \text{InfoMax}(D_n^{(i)}, U^{(i)})
\end{equation}

This term regularizes the task-specific experts, ensuring they learn specialized features that remain aligned with the global multimodal context.

\subsubsection{Performance-Adaptive Training Objective}
Training on heterogeneous tasks often leads to optimization imbalances. To address this, we design a performance-adaptive objective function. For task $n$, the base generation loss is the standard negative log-likelihood:
\begin{equation}
\mathcal{L}_{\text{gen}, n} = -\frac{1}{L} \sum_{t=1}^{L} \log P(\hat{w}_t \mid \hat{w}{<t}, \mathbf{T}, \mathbf{I}, \mathcal{A}_n)
\end{equation}

We introduce a dynamic weight $w_n$ that adapts based on the current task performance $a_n \in [0, 1]$ (e.g., F1 score):

\begin{equation}
w_n = (1 - a_n)^\gamma ,
\end{equation}

where $\gamma \ge 0$ is a focusing parameter. The total training objective $\mathcal{L}$ combines the weighted generation loss and the alignment term:

\begin{equation}
\mathcal{L} = \sum_{n} w_n \cdot \mathcal{L}_{\text{gen}, n} + \beta \cdot \mathcal{L}_{\text{align}}
\end{equation}

\subsubsection{Anchor-Aware Inference}
During inference, we employ an anchor-aware nucleus sampling strategy. While we sample tokens from the top-$p$ probability mass to ensure linguistic diversity, the generation process remains strictly conditioned on the semantic anchors $\mathcal{A}_n$ via the attention mechanism. This ensures that while the syntactic structure of the generated text varies, the core entities, relations, and event triggers remain invariant and consistent with the ground truth.

\begin{figure}[t]
  \centering
    \includegraphics[width=1\linewidth]{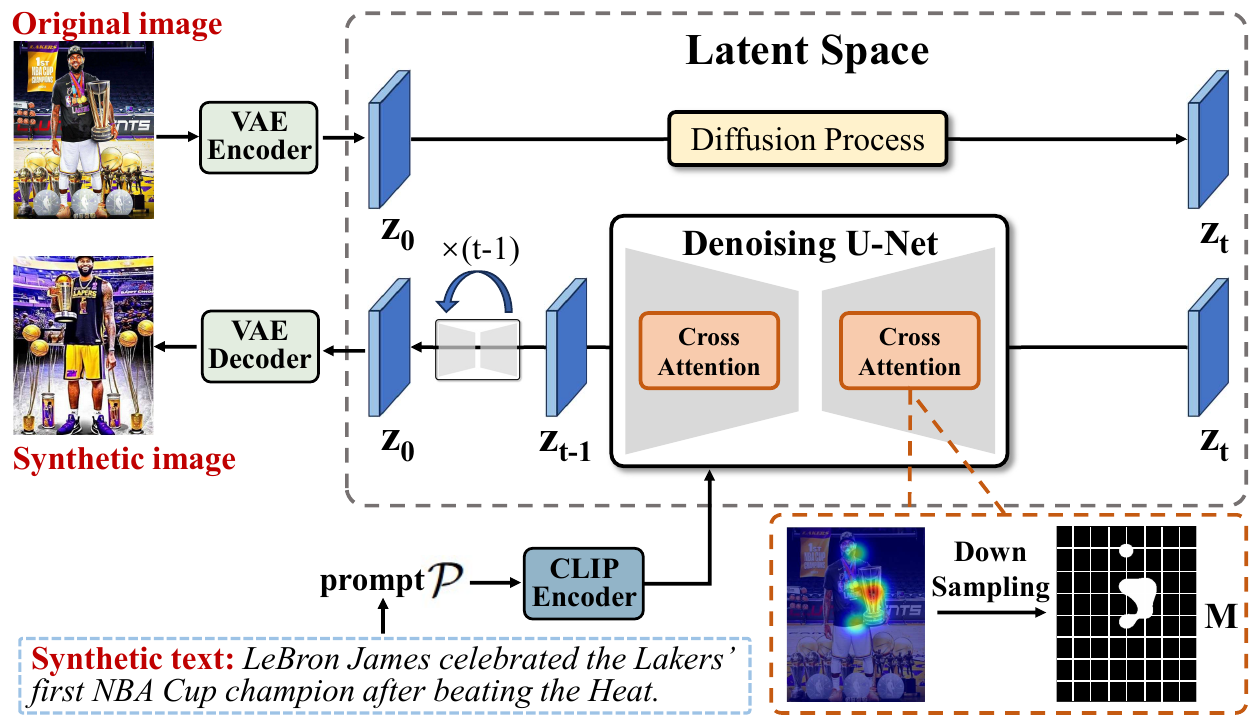}
\caption{Illustration of the Anchor-Preserving Image Synthesis. The process begins by constructing an \textbf{Anchor-Weighted Prompt} that combines the synthetic text with the explicit semantic anchors. Simultaneously, the Original Image is encoded into a latent representation $\mathbf{z}_0$ via a VAE Encoder. During the iterative denoising steps, the diffusion model employs a binary preservation mask $\mathbf{M}$ to enforce a dual constraint: regions corresponding to anchors are retained directly from the original latent $\mathbf{z}_0$ to ensure key entity identity preservation, while non-anchor (background) regions are synthesized by the denoiser conditioned on the prompt to introduce diversity, resulting in the final Synthetic Image.}
  \label{fig:framework1}
\end{figure}

\subsection{Anchor-Preserving Image Synthesis}
While standard augmentation methods often suffer from semantic drift—where the visual identity of key entities is altered—SAMA employs an Anchor-Preserving Diffusion strategy. We utilize a Latent Diffusion Model (LDM) in an image-to-image setting with two critical enhancements to enforce alignment with the semantic anchors $\mathcal{A}$.

\subsubsection{Anchor-Weighted Prompt Construction}
Instead of using the raw generated sentence, which may dilute the focus on key entities, we construct a weighted structural prompt $\mathcal{P}$. We assign higher attention weights to tokens corresponding to the Entity Anchors to ensure their visual features take precedence:
\begin{equation}
\mathcal{P} = \text{``A photo of } (\mathcal{A}{\text{entity}}:\omega) \text{ in } \mathcal{A}{\text{event}} \text{ context''} ,
\end{equation}

where $\omega > 1.0$ is an emphasis weight that forces the CLIP text encoder to prioritize the semantic anchors.

\subsubsection{Spatially-Constrained Denoising}
To preserve the visual integrity of entities while diversifying the context, we introduce a masked latent blending mechanism.First, the original image $\mathbf{I}$ is encoded into the latent space $\mathbf{z}_0$. We derive a binary preservation mask $\mathbf{M}$ from the cross-attention maps of the anchor tokens, where $\mathbf{M}_{ij}=1$ denotes foreground anchor regions and $\mathbf{M}_{ij}=0$ denotes background. 

During the reverse diffusion process, at each denoising step $t \to t-1$, we blend the model's prediction with the noisy original image to enforce the spatial constraint:
\begin{equation}
\mathbf{z}_{t-1} = \mathbf{M} \odot \mathbf{z}_{t-1}^{\text{src}} + (1 - \mathbf{M}) \odot \mathbf{z}_{t-1}^{\text{gen}} ,
\end{equation}

where $\odot$ denotes element-wise multiplication. $\mathbf{z}_{t-1}^{\text{gen}}$ is the denoised latent predicted by the model conditioned on prompt $\mathcal{P}$, introducing diversity in the background. $\mathbf{z}_{t-1}^{\text{src}}$ is the forward-noised latent of the original image at timestep $t-1$, ensuring that the pixels corresponding to semantic anchors (e.g., ``LeBron James'') remain strictly faithful to the source identity.

\subsection{Label Projection and Dual-Constraint Filtering}
Since the generation of both text $\hat{\mathbf{T}}$ and image $\hat{\mathbf{I}}$ is strictly conditioned on the task-specific semantic anchors, the synthetic data naturally inherits the semantic structure of the original sample. Therefore, we employ a label projection strategy where the original ground-truth annotations $\mathbf{Y}$ are directly assigned to the synthesized pair $(\hat{\mathbf{T}}, \hat{\mathbf{I}})$, eliminating the need for external re-annotation.

To ensure the reliability of this projection against potential hallucinations, we implement a Dual-Constraint Filtering mechanism. For each training sample, we generate $K=5$ candidates and select the optimal one based on a confidence score $S_{\text{conf}}$:

\subsubsection{Dual-Constraint Scoring Metric}
The score balances cross-modal alignment with anchor fidelity:
\begin{equation}
S_{\text{conf}} = \alpha \cdot \underbrace{\text{Sim}_{\text{CLIP}}(\hat{\mathbf{T}}, \hat{\mathbf{I}})}_{\text{Cross-Modal Alignment}} + (1 - \alpha) \cdot \underbrace{\text{Sim}_{\text{Sem}}(\hat{\mathbf{T}}, \mathcal{A})}_{\text{Anchor Fidelity}} ,
\end{equation}
where $\text{Sim}_{\text{CLIP}}$ measures the cosine similarity between the generated text and image embeddings, ensuring visual-textual consistency. $\text{Sim}_{\text{Sem}}$ measures the semantic similarity between the generated text and the structured anchor string. This penalty term filters out samples where the model fails to include the mandatory entities or relations defined in the anchors.

\subsubsection{Optimal Sample Selection}
We employ a maximum-confidence selection strategy. Candidates falling below a consistency threshold $\tau = 0.75$ are discarded. From the remaining pool, we select the highest-scoring pair for inclusion in the augmented dataset $\mathcal{D}_{\text{syn}}$:
\begin{equation}
    (\hat{\mathbf{T}}^*, \hat{\mathbf{I}}^*) = \operatorname*{argmax}_{k \in \{1 \dots K\}} S_{\text{conf}}^{(k)}
\end{equation}
This rigorous process ensures that SAMA generates data that is not only diverse but also semantically precise and valid for training.

\section{Experiments}
\begin{table}[h]
\centering
\setlength{\tabcolsep}{0.8mm}
\renewcommand{\arraystretch}{1.1}  
\caption{Statistics of the datasets used in our experiments.}
\label{tab:dataset_stats}
\scalebox{0.98}{
\begin{tabular}{llcccc}
\toprule
\textbf{Task} & \textbf{Dataset} & \textbf{Train} & \textbf{Dev} & \textbf{Test} & \textbf{Num of Types} \\
\midrule
MNER & Twitter-15 & 4,000  & 1,000 & 3,257  & 4  \\ \hline
MRE  & MNRE-V2    & 12,247 & 923   & 832    & 23 \\
\hline
MED  & M$^2$E$^2$ & --     & --    & 309    & 8  \\
\hline
MEAE & M$^2$E$^2$ & --     & --    & 309    & 15 \\
\bottomrule
\end{tabular}
}
\end{table}

\begin{table}[h]
\centering
\caption{Hyperparameter settings for the proposed SAMA framework.}
\label{tab:hyperparams}
\resizebox{0.9\columnwidth}{!}{
\begin{tabular}{l|lcc}
\toprule
\textbf{Module} & \textbf{Parameter} & \textbf{Symbol} & \textbf{Value} \\
\midrule
\multirow{2}{*}{Optimization} & Focusing Factor & $\gamma$ & 2.0 \\
 & Harmonization Weight & $\beta$ & 0.1 \\
\midrule
\multirow{2}{*}{Generation} & Prompt Emphasis & $\omega$ & 1.2 \\
 & Nucleus Sampling & $p$ & 0.9 \\
\midrule
\multirow{2}{*}{Filtering} & Alignment Weight & $\alpha$ & 0.6 \\
 & Confidence Threshold & $\tau$ & 0.75 \\
\bottomrule
\end{tabular}
}
\end{table}

\begin{table*}[t]
\centering
\renewcommand{\arraystretch}{1}
\caption{Performance comparison among different augmentation methods on Twitter-15 for the MNER task, MNRE for MRE and $\mathrm{M^2E^2}$ for the MEE task in low-resource settings and full-supervision settings.}
\label{tab:main_results}
\begin{tabular}{llcccccccccccc}
\hline
\multicolumn{1}{c}{\multirow{2}{*}{\textbf{Datasets}}} &
  \multirow{2}{*}{\textbf{Methods}} &
  \multicolumn{3}{c}{\textbf{10\%}} &
  \multicolumn{3}{c}{\textbf{20\%}} &
  \multicolumn{3}{c}{\textbf{40\%}} &
  \multicolumn{3}{c}{\textbf{100\%}} \\
\multicolumn{1}{c}{} &
   &
  \textbf{Pre.} &
  \textbf{Rec.} &
  \textbf{F1} &
  \textbf{Pre.} &
  \textbf{Rec.} &
  \textbf{F1} &
  \textbf{Pre.} &
  \textbf{Rec.} &
  \textbf{F1} &
  \textbf{Pre.} &
  \textbf{Rec.} &
  \textbf{F1} \\ \hline

 & 
  MMT5 &
  63.9 & 63.9 & 63.9 & 68.7 & 70.9 & 69.8 & 73.2 & 75.4 & 74.3 & 74.6 & 78.1 & 76.3 \\
 &
  \hspace{2em}-w/ mixGen &
  68.2 & 63.7 & 65.9 & 70.5 & 71.7 & 71.1 & 73.0 & 75.5 & 74.2 & 74.5 & 77.7 & 76.1 \\
 &
  \hspace{2em}-w/ GMDA &
  69.6 & 67.3 & 68.4 & 72.9 & 73.0 & 73.0 & 73.9 & 76.4 & 75.1 & 75.8 & \textbf{78.9} & 77.3 \\
\rowcolor{lightgray} \cellcolor{white} &
  \hspace{2em}-w/ SAMA(ours) &
  \textbf{71.6} & \textbf{68.9} & \textbf{70.2} & \textbf{75.3} & \textbf{73.8} & \textbf{74.5} & \textbf{76.8} & \textbf{78.6} & \textbf{77.7} & \textbf{76.7} & 78.5 & \textbf{77.6} \\ \cline{2-14} 
 &
  PGIM &
  72.6 & 71.7 & 72.2 & 73.2 & 77.1 & 75.1 & 75.2 & 78.3 & 76.7 & 77.3 & 80.2 & 78.7 \\
 &
  \hspace{2em}-w/ mixGen &
  71.4 & 74.1 & 72.7 & 75.4 & 77.2 & 76.3 & 76.7 & 77.9 & 77.3 & 77.2 & 78.9 & 78.0 \\
 &
  \hspace{2em}-w/ GMDA &
  73.2 & 74.6 & 73.9 & 75.9 & 76.5 & 76.2 & 76.8 & 78.4 & 77.6 & \textbf{78.3} & 80.1 & 79.2 \\
\rowcolor{lightgray} \cellcolor{white} \multirow{-8}{*}{\textbf{Twitter-15}} &
  \hspace{2em}-w/ SAMA(ours) &
  \textbf{75.8} & \textbf{75.3} & \textbf{75.6} & \textbf{76.7} & \textbf{78.6} & \textbf{77.6} & \textbf{77.0} & \textbf{79.6} & \textbf{78.3} & 78.2 & \textbf{80.9} & \textbf{79.5} \\ \hline

 &
  HVPNeT &
  75.1 & 76.5 & 75.8 & 78.4 & 78.5 & 78.5 & 80.9 & 79.4 & 80.2 & 83.6 & 80.8 & 81.9 \\
 &
  \hspace{2em}-w/ mixGen &
  75.4 & 76.1 & 75.7 & 78.6 & 78.8 & 78.7 & 80.3 & \textbf{80.7} & 80.5 & 83.8 & 81.5 & 82.6 \\
 &
  \hspace{2em}-w/ CAMIM &
  76.5 & 76.0 & 76.2 & 79.3 & 79.1 & 79.2 & 79.8 & 80.2 & 80.0 & 84.0 & \textbf{82.2} & 83.1 \\
\rowcolor{lightgray} \cellcolor{white} & 
  \hspace{2em}-w/ SAMA(ours) &
  \textbf{78.8} & \textbf{77.6} & \textbf{78.2} & \textbf{82.0} & \textbf{81.7} & \textbf{81.9} & \textbf{81.3} & 80.5 & \textbf{80.9} & \textbf{84.9} & 82.0 & \textbf{83.4} \\ \cline{2-14} 
 &
  TMR &
  82.7 & 82.1 & 82.4 & 85.6 & 83.9 & 84.7 & 87.6 & 85.3 & 86.4 & 90.5 & 87.7 & 89.1 \\
 &
  \hspace{2em}-w/ mixGen &
  84.5 & 82.8 & 83.6 & 86.5 & 85.3 & 85.9 & 87.9 & 86.3 & 87.1 & 90.4 & 88.1 & 89.2 \\
 &
  \hspace{2em}-w/ CAMIM &
  84.7 & 83.2 & 83.9 & 86.9 & \textbf{85.9} & \textbf{86.4} & 88.5 & 86.0 & 87.2 & 91.0 & \textbf{88.9} & \textbf{90.0} \\
\rowcolor{lightgray} \cellcolor{white} \multirow{-8}{*}{\textbf{MNRE}} &
  \hspace{2em}-w/ SAMA(ours) &
  \textbf{85.3} & \textbf{84.3} & \textbf{84.8} & \textbf{87.5} & 85.2 & {86.3} & \textbf{89.3} & \textbf{87.5} & \textbf{88.4} & \textbf{91.3} & 88.6 & 89.9 \\ \hline

 & 
  WASE &
  32.8 & 54.2 & 40.9 & 38.4 & 56.1 & 45.5 & 41.5 & 57.3 & 48.0 & 43.0 & 62.1 & 50.8 \\
 &
  \hspace{2em}-w/ mixGen &
  32.9 & 55.1 & 41.2 & 39.1 & 56.3 & 46.0 & 40.8 & 57.8 & 47.8 & 45.1 & 58.9 & 51.1 \\
 &
  \hspace{2em}-w/ CAMEL &
  34.1 & 54.9 & 42.1 & 39.7 & \textbf{58.5} & \textbf{47.7} & 44.9 & 58.3 & 50.6 & \textbf{49.8} & 60.2 & 53.7 \\
\rowcolor{lightgray} \cellcolor{white} & 
  \hspace{2em}-w/ SAMA(ours) &
  \textbf{34.3} & \textbf{56.8} & \textbf{42.8} & \textbf{40.3} & 57.8 & 47.5 & \textbf{45.5} & \textbf{58.4} & \textbf{51.2} & 48.6 & \textbf{63.6} & \textbf{55.1} \\ \cline{2-14} 
 &
  UniCL &
  34.5 & 55.3 & 42.5 & 38.9 & 57.1 & 46.3 & 42.2 & 60.9 & 50.6 & 44.1 & \textbf{67.7} & 53.4 \\
 &
  \hspace{2em}-w/ mixGen &
  35.4 & \textbf{56.2} & 43.4 & 38.7 & 58.1 & 46.5 & 44.5 & 59.2 & 51.2 & 46.3 & 67.5 & 54.9 \\
 &
  \hspace{2em}-w/ CAMEL &
  39.9 & 56.1 & 46.6 & 42.8 & \textbf{63.5} & 51.2 & 46.5 & 64.3 & 54.2 & 49.2 & 66.7 & 54.7 \\
\rowcolor{lightgray} \cellcolor{white} \multirow{-8}{*}{\textbf{{$\mathrm{M^2E^2}$} MED}} &
  \hspace{2em}-w/ SAMA(ours) &
  \textbf{42.8} & 55.6 & \textbf{48.4} & \textbf{45.7} & 62.4 & \textbf{52.8} & \textbf{49.6} & \textbf{64.9} & \textbf{56.2} & \textbf{53.2} & 65.7 & \textbf{58.8} \\ \hline

 & 
  WASE &
  13.8 & 13.5 & 13.7 & 14.9 & 15.3 & 15.1 & 16.1 & 16.3 & 16.2 & 19.5 & 18.9 & 19.2 \\
 &
  \hspace{2em}-w/ mixGen &
  14.5 & 14.1 & 14.3 & 15.8 & 16.2 & 16.0 & 17.2 & 16.8 & 17.0 & 19.4 & 19.5 & 19.4 \\
 &
  \hspace{2em}-w/ CAMEL &
  15.8 & \textbf{17.1} & 16.4 & 17.6 & 17.4 & 17.5 & 19.5 & 20.6 & 20.0 & 22.2 & \textbf{23.8} & 23.0 \\
\rowcolor{lightgray} \cellcolor{white} & 
  \hspace{2em}-w/ SAMA(ours) &
  \textbf{17.3} & 16.7 & \textbf{17.0} & \textbf{19.5} & \textbf{20.0} & \textbf{19.8} & \textbf{21.5} & \textbf{21.9} & \textbf{21.7} & \textbf{23.5} & 23.3 & \textbf{23.4} \\ \cline{2-14} 
 &
  UniCL &
  15.8 & 16.1 & 16.0 & 18.2 & 18.4 & 18.3 & 20.5 & 20.1 & 20.3 & 24.3 & 22.6 & 23.4 \\
 &
  \hspace{2em}-w/ mixGen &
  18.5 & 18.2 & 18.4 & 20.7 & 20.1 & 20.4 & 23.1 & 23.8 & 23.4 & 27.6 & 24.9 & 26.2 \\
 &
  \hspace{2em}-w/ CAMEL &
  19.1 & \textbf{20.7} & 19.9 & 20.5 & \textbf{22.5} & 21.5 & 25.9 & 24.1 & 24.9 & 28.5 & 27.3 & 27.9 \\
\rowcolor{lightgray} \cellcolor{white} \multirow{-8}{*}{\textbf{{$\mathrm{M^2E^2}$} MEAE}} &
  \hspace{2em}-w/ SAMA(ours) &
  \textbf{20.7} & 20.6 & \textbf{20.7} & \textbf{24.9} & 22.0 & \textbf{23.4} & \textbf{26.8} & \textbf{25.3} & \textbf{26.0} & \textbf{30.7} & \textbf{29.1} & \textbf{29.9} \\ \hline
\end{tabular}
\end{table*}

\subsection{Experimental Settings}
\subsubsection{Datasets}
We evaluate SAMA on three benchmark datasets that represent the core tasks of MIE:
\begin{itemize}
\item \textbf{MNER}: We use Twitter-15~\cite{zhang2018adaptive}, a widely used dataset curated from social media containing multimodal tweets annotated with entity types.
\item \textbf{MRE}: We adopt MNRE~\cite{zheng2021multimodal}, a prominent benchmark for multimodal relation extraction derived from social media posts.
\item \textbf{MEE}: Following standard protocols~\cite{li2020cross,liu2022multimedia}, we evaluate on the $\mathrm{M^2E^2}$~\cite{li2020cross} benchmark. Since $\mathrm{M^2E^2}$ lacks a dedicated training set, we construct the training data using ACE2005~\cite{walker2006ace} (text-only) and imSitu~\cite{yatskar2016situation} (image-only). Specifically, we map the 98 activity verbs from imSitu and the 33 event types from ACE2005 to the 8 target event types of $\mathrm{M^2E^2}$, creating a heterogeneous training environment that tests SAMA's ability to align cross-modal semantics.
\end{itemize}

Detailed statistics for all datasets are provided in Table~\ref{tab:dataset_stats}. To rigorously assess robustness in data-scarce scenarios, we construct three low-resource splits by randomly sampling 10\%, 20\%, and 40\% of the training and validation data, while keeping the full test set for consistent evaluation.

\subsubsection{Implementation Details}
Our SAMA framework is built upon the InstructBLIP architecture~\cite{dai2023instructblip}, utilizing FlanT5-XL (3B)~\cite{chung2024scaling} as the LLM backbone and ViT-g/14~\cite{fang2023eva} as the visual encoder. For the Collaborative Multi-Experts mechanism, we implement both the Universal Adapter $U(x)$ and Task-specific Adapters $D_n(x)$ using LoRA~\cite{hulora}. To balance capacity and efficiency, we set the LoRA rank $r=8$ and dropout to 0.1 for all adapters. The model is optimized using AdamW with a batch size of 2 and a learning rate of 5e-5. The hyperparameters are shown in Table~\ref{tab:hyperparams}.

For image synthesis, we employ Stable Diffusion v1.5~\cite{rombach2022high} with a denoising strength of 0.8 and a guidance scale of 10. All experiments are conducted on 4 NVIDIA Tesla A800 GPUs.

\subsubsection{Evaluation Metrics}
We report Precision (Pre.), Recall (Rec.), and F1 Score (F1) for all tasks. Predictions are considered correct only if they strictly match the ground-truth entity spans, types, or relation/event schemas.

\subsection{Comparison Systems}
To validate the efficacy of SAMA, we adopt a ``Plug-and-Play" evaluation strategy. We select representative state-of-the-art models for each task as Backbones and compare SAMA against other DA methods when applied to these backbones.

\subsubsection{Task Backbones}
We employ the following competitive models to benchmark the improvement brought by data augmentation:
\begin{itemize}
\item \textbf{For MNER}: We use MMT5~\cite{wang2023fine}, a generative paraphrase model, and PGIM~\cite{li2023prompting}, the current SOTA method that leverages ChatGPT for auxiliary knowledge knowledge.
\item \textbf{For MRE}: We adopt HVPNeT~\cite{chen2022good}, which uses visual prefixes for error-correction, and TMR~\cite{zheng2023rethinking}, which employs multimodal back-translation.
\item \textbf{For MEE}: We use the object-detection variant of WASE (\textbf{WASE\textsubscript{obj}})~\cite{li2020cross} and UniCL~\cite{liu2022multimedia}, the SOTA method on $\mathrm{M^2E^2}$ utilizing contrastive learning.
\end{itemize}
\subsubsection{Data Augmentation Baselines}
We compare SAMA with four augmentation strategies:
\begin{itemize}
\item \textbf{MixGen}~\cite{hao2023mixgen}: A task-agnostic baseline that linearly interpolates images and text.
\item \textbf{GMDA}~\cite{li2024generative}: A generative framework specifically designed for MNER augmentation.
\item \textbf{CAMIM}~\cite{zhang2024caption}: A caption-aware augmentation method tailored for MNRE.
\item \textbf{CAMEL}~\cite{du2023training}: A generative augmentation pipeline designed for MEE.
\end{itemize}

Comparison with these methods allows us to demonstrate SAMA's superiority over both general-purpose augmentation (MixGen) and task-specific approaches (GMDA, CAMIM, CAMEL).

\subsection{Results in Low-Resource Settings} Table~\ref{tab:main_results} presents the comparative performance of SAMA against both task-agnostic and task-specific DA methods across three low-resource splits.

\subsubsection{Performance on MNER} SAMA consistently outperforms all baselines. Notably, in the extreme low-resource setting (10\%), SAMA achieves a significant margin over the task-specific SOTA, GMDA, improving the F1 score by 1.7\% when applied to the PGIM backbone. While PGIM leverages ChatGPT for external knowledge, SAMA complements this by providing high-fidelity, in-domain visual-textual pairs, proving that structured synthetic data offers value beyond pure textual knowledge retrieval.

\subsubsection{Performance on MRE} Similar trends are observed in the MRE task. SAMA demonstrates superior robustness compared to CAMIM, the relation-oriented augmentation baseline. Specifically, on the MNRE dataset (10\% split), SAMA boosts the HVPNeT baseline by approximately 2.0\% F1. Crucially, MixGen often yields negligible or even negative gains due to the noise introduced by linear interpolation. In contrast, SAMA's anchor-driven generation preserves the semantic logic of relations (e.g., ``MemberOf"), ensuring that synthetic samples remain valid training signals.

\subsubsection{Performance on MEE} For the complex MEE task, SAMA significantly surpasses the unimodal augmentation method CAMEL. We observe the largest performance boost in the 10\% setting, outperforming the prior SOTA by 5.9\% F1 on MED and 4.7\% F1 on MEAE. This suggests that our \textit{Collaborative Multi-Experts} mechanism effectively transfers event-level structural knowledge (arguments and triggers) from the LLM to the extraction model, bridging the gap that purely visual or textual augmentation cannot address.

\subsection{Results in Full-Supervision Settings}
Even with abundant training data, SAMA provides consistent performance gains, addressing the long-tail distribution often present in full datasets. As shown in Table~\ref{tab:main_results}, SAMA-augmented TMR achieves a new state-of-the-art F1 of 77.6\% on Twitter-15, surpassing MMT5. Similarly, for MED, SAMA pushes UniCL to 58.8\%, a 4.1\% improvement over CAMEL. This indicates that SAMA does not merely replicate existing patterns but introduces necessary semantic diversity (e.g., varied backgrounds for the same entity) via its \textit{Anchor-Preserving Diffusion}, thereby improving the model's generalization to unseen test samples.

\subsection{In-Depth Analysis}

{
\color{red}
\begin{table*}[t]
\centering
\renewcommand{\arraystretch}{1}
\caption{Comprehensive Ablation Study of SAMA across all benchmark datasets under different data ratios (10\%, 20\%, 40\%, and 100\%).}
\label{tab:comprehensive_ablation}
\resizebox{\textwidth}{!}{
\begin{tabular}{llcccccccccccc}
\hline
\multicolumn{1}{c}{\multirow{2}{*}{\textbf{Datasets}}} &
  \multirow{2}{*}{\textbf{Methods (Ablation Variants)}} &
  \multicolumn{3}{c}{\textbf{10\%}} &
  \multicolumn{3}{c}{\textbf{20\%}} &
  \multicolumn{3}{c}{\textbf{40\%}} &
  \multicolumn{3}{c}{\textbf{100\%}} \\
\multicolumn{1}{c}{} &
   &
  \textbf{Pre.} & \textbf{Rec.} & \textbf{F1} &
  \textbf{Pre.} & \textbf{Rec.} & \textbf{F1} &
  \textbf{Pre.} & \textbf{Rec.} & \textbf{F1} &
  \textbf{Pre.} & \textbf{Rec.} & \textbf{F1} \\ \hline

 \rowcolor{lightgray} \cellcolor{white}& 
  \textbf{Base Model with SAMA} &
  \textbf{75.8} & \textbf{75.3} & \textbf{75.6} & \textbf{76.7} & \textbf{78.6} & \textbf{77.6} & \textbf{77.0} & \textbf{79.6} & \textbf{78.3} & \textbf{78.2} & \textbf{80.9} & \textbf{79.5} \\
 &
  \hspace{2em}- w/o Universal Expert &
  72.4 & 72.0 & 72.2 & 74.8 & 75.4 & 75.1 & 75.5 & 77.6 & 76.5 & 77.1 & 79.6 & 78.3 \\
 &
  \hspace{2em}- w/o Knowledge Harmonization &
  70.1 & 69.8 & 69.9 & 73.5 & 74.6 & 74.0 & 74.9 & 76.8 & 75.8 & 76.8 & 79.1 & 77.9 \\
 &
  \hspace{2em}- w/o Univ. Exp. \& Harmonization &
  68.5 & 67.8 & 68.1 & 72.1 & 72.9 & 72.5 & 73.7 & 75.6 & 74.6 & 75.9 & 78.2 & 77.0 \\
 &
  \hspace{2em}- w/o Anchor-Preserving Diffusion &
  73.0 & 72.5 & 72.7 & 75.2 & 76.1 & 75.6 & 76.1 & 77.8 & 76.9 & 77.3 & 79.8 & 78.5 \\
\multirow{-6}{*}{\textbf{Twitter-15}} &
  \hspace{2em}- w/o Dual-Constraint Filtering &
  74.0 & 73.8 & 73.9 & 75.8 & 76.7 & 76.2 & 76.4 & 78.5 & 77.4 & 77.8 & 80.3 & 79.0 \\ \hline

 \rowcolor{lightgray} \cellcolor{white}&
  \textbf{Base Model with SAMA} &
  \textbf{85.3} & \textbf{84.3} & \textbf{84.8} & \textbf{87.5} & \textbf{85.2} & \textbf{86.3} & \textbf{89.3} & \textbf{87.5} & \textbf{88.4} & \textbf{91.3} & \textbf{88.6} & \textbf{89.9} \\
 &
  \hspace{2em}- w/o Universal Expert &
  82.0 & 81.5 & 81.8 & 85.5 & 83.0 & 84.2 & 87.8 & 85.8 & 86.8 & 90.2 & 87.3 & 88.7 \\
 &
  \hspace{2em}- w/o Knowledge Harmonization &
  79.5 & 78.8 & 79.0 & 84.3 & 82.0 & 83.1 & 86.9 & 85.0 & 85.9 & 89.7 & 86.8 & 88.2 \\
 &
  \hspace{2em}- w/o Univ. Exp. \& Harmonization &
  78.2 & 77.5 & 77.8 & 82.8 & 80.3 & 81.5 & 85.5 & 83.6 & 84.5 & 89.0 & 86.1 & 87.5 \\
 &
  \hspace{2em}- w/o Anchor-Preserving Diffusion &
  80.8 & 80.1 & 80.5 & 85.8 & 83.5 & 84.6 & 88.0 & 86.1 & 87.0 & 90.4 & 87.5 & 88.9 \\
\multirow{-6}{*}{\textbf{MNRE}} &
  \hspace{2em}- w/o Dual-Constraint Filtering &
  83.5 & 83.0 & 83.3 & 86.6 & 84.3 & 85.4 & 88.6 & 86.8 & 87.7 & 90.8 & 88.1 & 89.4 \\ \hline

 \rowcolor{lightgray} \cellcolor{white}&
  \textbf{Base Model with SAMA} &
  \textbf{42.8} & \textbf{55.6} & \textbf{48.4} & \textbf{45.7} & \textbf{62.4} & \textbf{52.8} & \textbf{49.6} & \textbf{64.9} & \textbf{56.2} & \textbf{53.2} & \textbf{65.7} & \textbf{58.8} \\
 &
  \hspace{2em}- w/o Universal Expert &
  39.8 & 53.0 & 45.6 & 43.2 & 60.5 & 50.4 & 47.8 & 63.5 & 54.5 & 52.0 & 64.8 & 57.7 \\
 &
  \hspace{2em}- w/o Knowledge Harmonization &
  37.0 & 50.5 & 42.0 & 41.5 & 59.2 & 48.8 & 46.9 & 62.6 & 53.6 & 51.5 & 64.4 & 57.2 \\
 &
  \hspace{2em}- w/o Univ. Exp. \& Harmonization &
  35.8 & 49.2 & 41.4 & 39.3 & 56.8 & 46.5 & 45.0 & 61.0 & 51.8 & 50.6 & 63.5 & 56.3 \\
 &
  \hspace{2em}- w/o Anchor-Preserving Diffusion &
  40.5 & 53.8 & 46.8 & 44.0 & 61.3 & 51.2 & 48.3 & 63.8 & 55.0 & 52.3 & 65.1 & 58.0 \\
\multirow{-6}{*}{\textbf{{$\mathrm{M^2E^2}$} MED}} &
  \hspace{2em}- w/o Dual-Constraint Filtering &
  41.8 & 55.0 & 47.4 & 44.8 & 61.8 & 51.9 & 48.9 & 64.2 & 55.5 & 52.8 & 65.3 & 58.4 \\ \hline

 \rowcolor{lightgray} \cellcolor{white}&
  \textbf{Base Model with SAMA} &
  \textbf{20.7} & \textbf{20.6} & \textbf{20.7} & \textbf{24.9} & \textbf{22.0} & \textbf{23.4} & \textbf{26.8} & \textbf{25.3} & \textbf{26.0} & \textbf{30.7} & \textbf{29.1} & \textbf{29.9} \\
 &
  \hspace{2em}- w/o Universal Expert &
  18.5 & 18.0 & 18.3 & 22.8 & 20.4 & 21.5 & 25.3 & 24.0 & 24.6 & 29.5 & 28.1 & 28.8 \\
 &
  \hspace{2em}- w/o Knowledge Harmonization &
  17.1 & 16.8 & 16.9 & 21.7 & 19.3 & 20.4 & 24.6 & 23.3 & 23.9 & 29.1 & 27.7 & 28.4 \\
 &
  \hspace{2em}- w/o Univ. Exp. \& Harmonization &
  15.5 & 15.0 & 15.2 & 20.0 & 17.8 & 18.8 & 23.2 & 21.9 & 22.5 & 28.3 & 27.0 & 27.6 \\
 &
  \hspace{2em}- w/o Anchor-Preserving Diffusion &
  18.8 & 18.2 & 18.5 & 23.4 & 20.8 & 22.0 & 25.8 & 24.5 & 25.1 & 29.8 & 28.4 & 29.1 \\
\multirow{-6}{*}{\textbf{{$\mathrm{M^2E^2}$} MEAE}} &
  \hspace{2em}- w/o Dual-Constraint Filtering &
  20.0 & 19.6 & 19.8 & 24.2 & 21.5 & 22.8 & 26.3 & 25.0 & 25.6 & 30.2 & 28.8 & 29.5 \\ \hline
\end{tabular}
}
\end{table*}
}

\subsubsection{Ablation Study}
To rigorously quantify the individual contributions of the proposed modules and to substantiate their necessity in both data-scarce and data-rich regimes, we extend our ablation analysis to a more comprehensive experimental setting. Table~\ref{tab:comprehensive_ablation} reports the performance of all model variants on the four datasets under different training data ratios. In particular, we compare the complete SAMA framework against five ablated variants:

\begin{itemize}
\item \textbf{w/o Universal Expert}: Removing $U(x)$ and using only task-specific adapters.
\item \textbf{w/o Knowledge Harmonization}: Removing the alignment loss $\mathcal{L}_{\text{align}}$.

\item \textbf{w/o Univ. Exp. \& Harmonization}: Jointly removing both the universal expert and the harmonization mechanism to test synergistic effects.

\item \textbf{w/o Anchor-Preserving Diffusion}: Replacing our image synthesis with standard text-to-image generation (i.e., no image conditioning).
\item \textbf{w/o Dual-Constraint Filtering}: Removing the filtering step and using raw generated pairs.
\end{itemize}

As presented in Table~\ref{tab:comprehensive_ablation}, each component is essential and collectively gives rise to several salient observations:

\begin{enumerate}
\item \textbf{Impact of Collaborative Experts}: Removing either the \textit{Universal Adapter} or the \textit{Knowledge Harmonization} leads to a notable drop across all tasks. This confirms that shared semantic knowledge is crucial for complex understanding, and the teacher-student mechanism effectively regularizes the task-specific experts.
\item \textbf{Synergistic Effect of Joint Removal}: Crucially, the simultaneous removal of both the Universal Expert and Knowledge Harmonization mechanism results in a catastrophic performance degradation. This substantial decline—exceeding the individual drops of either component—demonstrates that the harmonization loss is not merely a regularizer but acts as a critical synergistic bridge. It ensures that the task-specific experts do not become isolated and overfitted, but instead remain anchored to the general multimodal semantic space provided by the universal expert.
\item \textbf{Impact under Different Data Ratios}: By tracking the ablation performance from 10\% to 100\% supervision, we observe a consistent trend. While the relative performance gap between the full SAMA framework and the ablated variants gradually narrows as task-specific data becomes abundant, the shared modules still provide a statistically significant improvement even under full supervision. This demonstrates that SAMA is not merely a low-resource remedy; its shared semantics effectively prevent task-specific adapters from overfitting to the idiosyncrasies of the training set, acting as a robust regularizer in high-resource scenarios.
\item \textbf{Impact of Visual Fidelity}: The \textit{w/o Anchor-Preserving} variant suffers a persistent decline across all settings. Without conditioning on the original image, standard diffusion models often hallucinate entity identities (e.g., generating a different person or background contex), breaking the alignment with ground-truth labels. Our spatial masking approach successfully mitigates this identity drift.
\item \textbf{Impact of Quality Control}: Removing \textit{Dual-Constraint Filtering} introduces noise into the augmentation pool, verifying that enforcing both cross-modal alignment and anchor fidelity is vital for training stability.
\end{enumerate}

\begin{figure*}[t]
\centering
\includegraphics[width=0.98\linewidth]{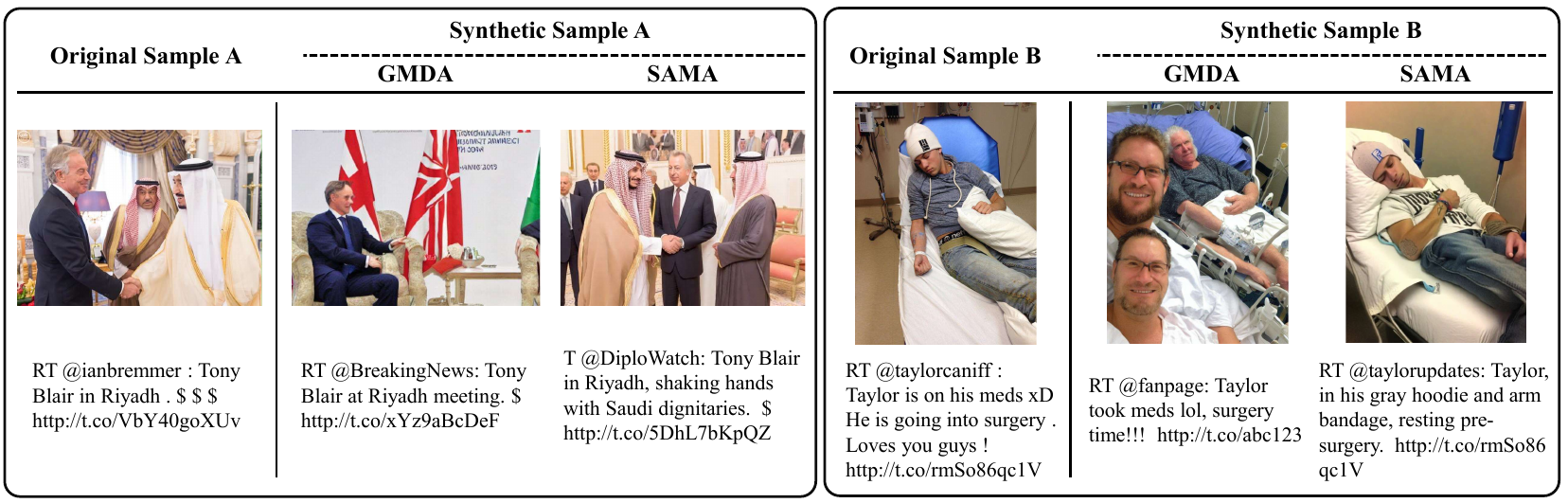}
\caption{Comparison between synthetic samples generated from GMDA and those generated from SAMA.}
\label{Fig:exp}
\end{figure*}

\begin{figure*}[htbp]
  \centering
  \includegraphics[width=0.99\linewidth]{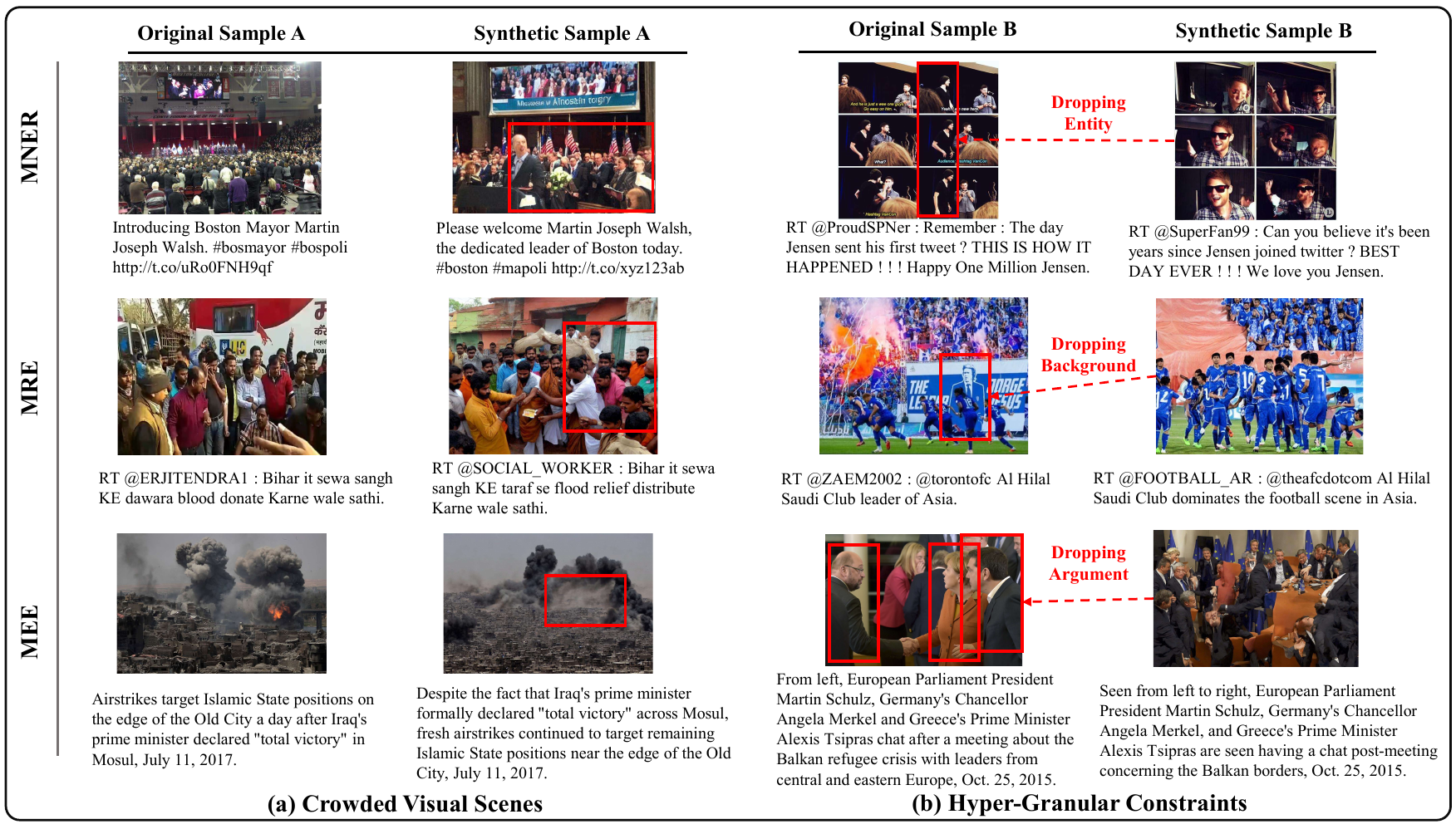} 
  \caption{Comprehensive visualization of SAMA failure cases across MNER, MRE, and MEE tasks. \textbf{(a) Crowded Visual Scenes (Left)}: In highly complex and dense backgrounds, inaccurate mask boundaries can lead to visual blending artifacts. \textbf{(b) Hyper-Granular Constraints (Right)}: When forced to handle an excessive number of semantic anchors, the model's attention may dilute, occasionally omitting minor constraints in the generated text.}
  \label{fig:errors}
\end{figure*}

\begin{figure*}[t]
  \centering
  \includegraphics[width=0.9\linewidth]{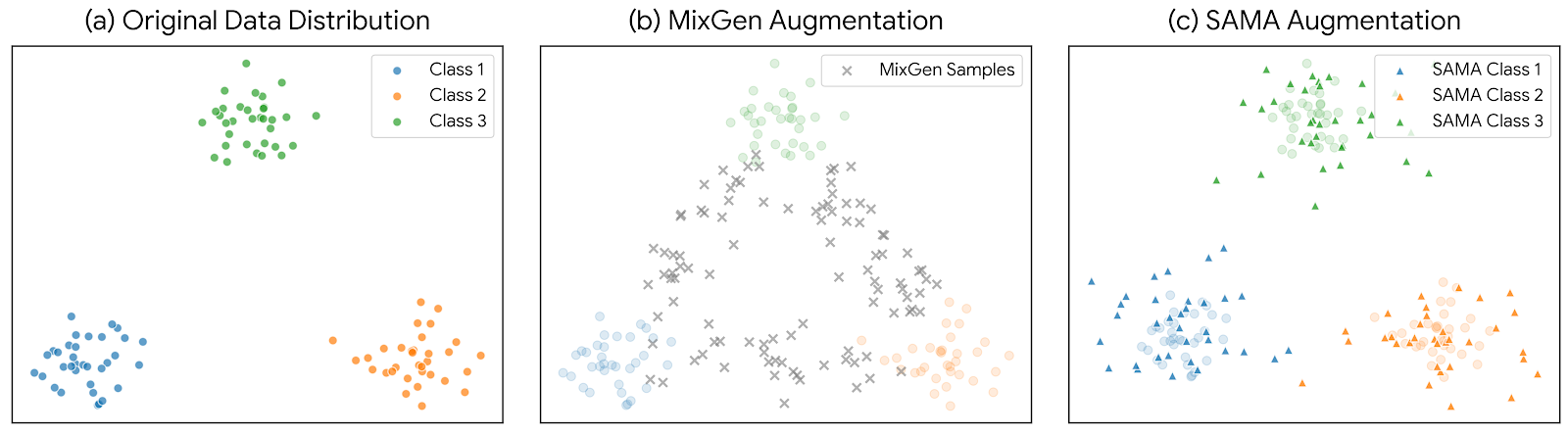}
  \caption{t-SNE visualization of feature distributions. (a) Original training data forms tight clusters. (b) MixGen generates samples between clusters (gray crosses), leading to semantic ambiguity. (c) SAMA produces diverse samples (colored triangles) that adhere strictly to the original semantic manifolds.}
  \label{fig:tsne}
\end{figure*}

\subsubsection{Case study}
Figure~\ref{Fig:exp} visualizes the qualitative superiority of SAMA over GMDA. For Sample A, GMDA suffers from semantic drift: it generates a generic ``handshake" image where the visual identity of the entity is lost. In stark contrast, SAMA's \textit{Anchor-Preserving Diffusion} strictly retains the facial features of ``Tony Blair" (the entity anchor) while successfully diversifying the context (e.g., enhancing the background flags and ceremonial attire). For Sample B, GMDA produces blurred, ambiguous equipment. SAMA, guided by structured anchors, synthesizes a sharper image with distinct medical markers (bandages, instruments), enriching the semantic context for the relation extraction model. These examples demonstrate SAMA's ability to generate \textit{high-fidelity, identity-preserving} augmentations, solving the hallucination issues prevalent in previous generative methods.

\subsubsection{Error Analysis and Limitations}
To transparently analyze the boundary conditions of our framework across different MIE tasks, we systematically visualize two primary failure modes in Figure~\ref{fig:errors}.

\begin{enumerate}
    \item \textbf{Crowded Visual Scenes} (Figure~\ref{fig:errors}a, Right Column): When the original image contains highly overlapping subjects—such as the dense audience in the MNER example or the clustered gathering in the MRE example—the cross-attention-derived preservation mask $\mathbf{M}$ may struggle to strictly delimit the subject boundaries. This occasionally results in minor \textit{blending artifacts} or ``ghosting'' along the edges of the preserved entities in the synthesized images (e.g., as highlighted by the red annotations in the synthetic samples).
    
    \item \textbf{Hyper-Granular Constraints} (Figure~\ref{fig:errors}b, Right Column): We also observe limitations in the model's token capacity when handling complex contextual generation. When an input sample is injected with an extremely dense set of semantic anchors (e.g., extracting comprehensive event arguments or multiple interactive relations from a single image), the MLLM's attention mechanism may dilute. This causes the synthesized image to unintentionally omit a minor visual or relational constraint (e.g., dropping a specific action or background entity, as marked in the synthetic text). Future work will explore coarse-to-fine iterative generation strategies to overcome this limitation.
\end{enumerate}

\subsubsection{Feature Space Visualization}
To intuitively understand the quality of the synthesized data, we visualize the feature distributions of the original training set, data augmented by MixGen, and data augmented by SAMA using t-SNE. As shown in Figure~\ref{fig:tsne}, the original data (a) forms distinct clusters. The data generated by MixGen (b), due to its linear interpolation mechanism, tends to populate the low-density regions between classes, causing ``manifold intrusion" and blurring decision boundaries. In contrast, SAMA (c) generates samples that are closely clustered around the original class centers but with expanded variance. This demonstrates that SAMA effectively introduces semantic diversity while preserving the high fidelity and discriminative power of the original features, avoiding the noise injection common in interpolation-based methods.

\begin{table}[t]
    \centering
    \caption{Comparison of generation quality on Twitter-15. 
    $\downarrow$ indicates lower is better, $\uparrow$ indicates higher is better.}
    \label{tab:quality_metrics}
    \setlength{\tabcolsep}{0.83mm}
    \renewcommand{\arraystretch}{1}
    \scalebox{0.99}{
    \begin{tabular}{lcccc}
        \toprule
        \multirow{2}{*}{\textbf{Method}} & \multicolumn{2}{c}{\textbf{Visual Quality}} & \multicolumn{2}{c}{\textbf{Textual Quality}} \\
        \cmidrule(lr){2-3} \cmidrule(lr){4-5}
        & \textbf{FID} ($\downarrow$) & \textbf{CLIP Score} ($\uparrow$) & \textbf{Diversity} ($\uparrow$) & \textbf{Perplexity} ($\downarrow$) \\
        \midrule
        MixGen     & 45.12 & 0.65 & 0.55 & 114.12 \\
        GMDA     & 31.67 & 0.72 & 0.89 & 104.96 \\
        \rowcolor{gray!10}
        \textbf{SAMA (Ours)}             & \textbf{24.35} & \textbf{0.78} & \textbf{0.95} & \textbf{98.45} \\
        \bottomrule
    \end{tabular}
    }
\end{table}

\subsubsection{Generation Quality Assessment}
To comprehensively evaluate the fidelity and alignment of the synthesized data, we assess both visual and textual quality using four standard metrics. Visual quality is measured by \textbf{Fréchet Inception Distance (FID)}~\cite{heusel2017gans} for realism and \textbf{CLIP Score}~\cite{radford2021learning} for cross-modal consistency. Following Li et al.~\cite{li2024generative}, we also evaluate textual quality using:
\begin{itemize}
    \item \textbf{Diversity}: Defined as the percentage of unique $n$-grams (where $n \in \{2,3,4\}$) in the generated sentences relative to the total number of $n$-grams. A higher score indicates richer linguistic variation.
    \item \textbf{Perplexity}\footnote{https://huggingface.co/docs/transformers/perplexity}: Calculated using a pre-trained GPT-2 model to measure the fluency and coherence of the generated text. Lower perplexity indicates more natural and grammatically correct sentences.
\end{itemize}

\begin{table*}[t]
\centering
\caption{Comparison of computational cost, training efficiency across different tasks, and inference overhead. \textbf{Gen. Time} denotes the offline time to generate one augmented sample. \textbf{Training Time} indicates the time required to train the downstream backbone model on the augmented dataset. ``--'' indicates the augmentation baseline is task-specific and inapplicable to that dataset.}
\label{tab:cost}
\setlength{\tabcolsep}{3.5mm}
\renewcommand{\arraystretch}{1}
\resizebox{\textwidth}{!}{
\begin{tabular}{lcccccc}
\toprule
\multirow{2}{*}{\textbf{Method}} & \textbf{Gen. Time} & \textbf{GPU Mem.} & \multicolumn{3}{c}{\textbf{Downstream Training Time (Hours)}} & \textbf{Inference} \\
\cmidrule(lr){4-6}
 & (s/sample) & (GB) & \textbf{MNER} (Twitter-15) & \textbf{MRE} (MNRE) & \textbf{MEE} ($\mathrm{M^2E^2}$) & \textbf{Overhead} \\
\midrule
MixGen  & $\approx$ 0.1 & 2.5  & $\approx$ 4.2 & $\approx$ 6.5 & $\approx$ 2.8 & \textbf{None} \\
GMDA & $\approx$ 1.5 & 12.0 & $\approx$ 4.2 & -- & -- & \textbf{None} \\
CAMIM& $\approx$ 1.8 & 13.5 & -- & $\approx$ 6.5 & -- & \textbf{None} \\
CAMEL & $\approx$ 2.0 & 14.0 & -- & -- & $\approx$ 2.8 & \textbf{None} \\
\rowcolor{gray!10}
\textbf{SAMA (Ours)}         & $\approx$ 3.5 & 18.5 & $\approx$ 4.2 & $\approx$ 6.5 & $\approx$ 2.8 & \textbf{None} \\
\bottomrule
\end{tabular}
}
\end{table*}

\begin{figure}[h]
  \centering
  \includegraphics[width=0.8\linewidth]{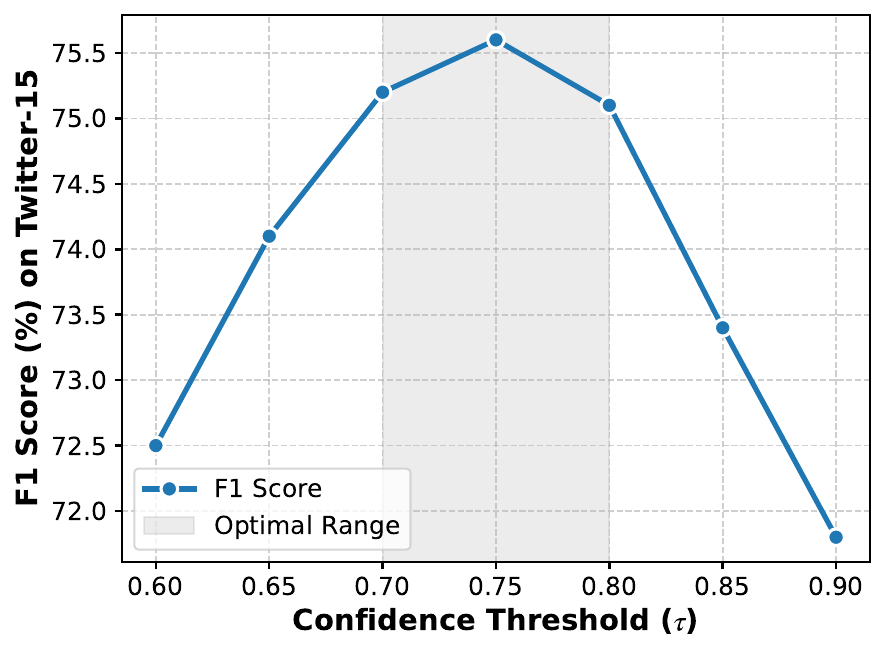}
  \caption{Sensitivity analysis of the confidence threshold $\tau$ on Twitter-15.}
  \label{fig:sensitivity}
\end{figure}

\begin{figure}[h]
  \centering
  \includegraphics[width=0.8\linewidth]{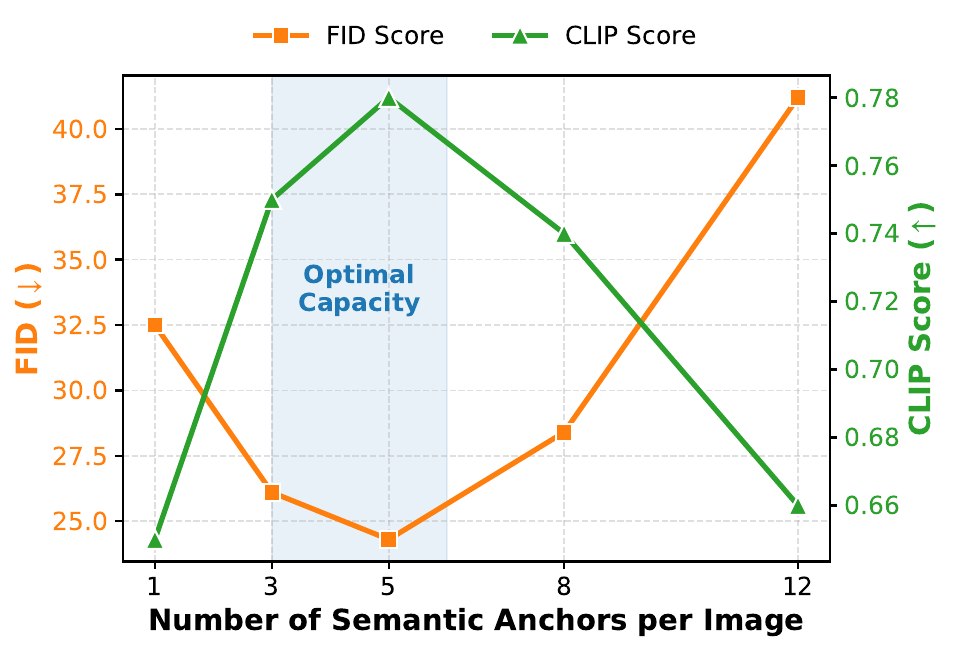}
  \caption{Impact of semantic anchor density on generation quality.}
  \label{fig:anchor_density}
\end{figure}

Table~\ref{tab:quality_metrics} presents the comparative results on the Twitter-15 dataset. 
Regarding \textbf{Visual Fidelity \& Alignment}, we observe that SAMA achieves the lowest FID and the highest CLIP Score, significantly outperforming both MixGen and GMDA. This confirms that our \textit{Anchor-Preserving Diffusion} strategy effectively maintains high visual realism while ensuring strict semantic alignment with the text. 

Simultaneously, for \textbf{Textual Diversity \& Fluency}, the results indicate that MixGen suffers from high perplexity due to the unnatural concatenation of sentences. While GMDA produces fluent text, its diversity is limited as it often generates repetitive patterns. In contrast, SAMA attains the highest Diversity score while maintaining excellent fluency (lowest Perplexity). This demonstrates that our SAMA framework successfully generates syntactically diverse yet semantically precise training samples.

\subsubsection{Computational Cost and Efficiency Analysis}
We explicitly evaluate the computational overhead of SAMA across all three tasks to address potential concerns regarding deployment efficiency. It is imperative to emphasize that SAMA operates entirely as an offline data augmentation pipeline.

As detailed in Table~\ref{tab:cost}, we compare the efficiency metrics against representative baselines: MixGen~\cite{hao2023mixgen}, GMDA~\cite{li2024generative}, CAMIM~\cite{zhang2024caption}, and CAMEL~\cite{du2023training}. We report the downstream training time for each task using its respective SOTA backbone: \textbf{PGIM} for MNER, \textbf{HVPNeT} for MRE, and \textbf{UniCL} for MEE. Notably, because GMDA, CAMIM, and CAMEL are rigidly designed for specific tasks, they cannot be applied across all datasets. In contrast, SAMA and the interpolation-based MixGen serve as unified, task-agnostic frameworks.

While SAMA naturally requires more offline generation time and memory compared to simpler methods like MixGen (due to the iterative Anchor-Preserving Diffusion process), these costs are incurred only once. Crucially, since the augmentation volume is fixed across all methods, the \textbf{downstream training time remains strictly consistent} for a given dataset. Most importantly, SAMA introduces \textbf{zero} additional latency, memory footprint, or parameter overhead to the downstream inference phase. Given the significant F1 performance gains across diverse tasks, this one-time offline computational cost is highly justified.

\subsubsection{Parameter Sensitivity Analysis}
To demonstrate the robustness of the Dual-Constraint Filtering mechanism, we conduct a sensitivity analysis on the confidence threshold $\tau$. As illustrated in Figure~\ref{fig:sensitivity}, we evaluate the F1 score on the Twitter-15 dataset (10\% ratio) across $\tau \in \{0.60, 0.65, 0.70, 0.75, 0.80, 0.85, 0.90\}$. Setting $\tau$ too low admits noisy samples with semantic drift, leading to suboptimal performance. Conversely, an aggressive threshold severely restricts the diversity of the augmented data pool, causing the model to overfit on a limited set of synthetic patterns. The performance remains robust and peaks when $\tau \in [0.70, 0.80]$, empirically justifying our selection of $\tau = 0.75$.

\subsubsection{Impact of Semantic Anchor Density}
We further investigate how the density and granularity of semantic anchors impact the generation quality on Twitter-15. We manipulate the number of injected entity/relation anchors per image and evaluate the resulting visual realism (FID) and cross-modal alignment (CLIP Score). As shown in Figure~\ref{fig:anchor_density}, we observe a clear ``Goldilocks effect." Providing too few anchors leads to unconstrained hallucinations, resulting in lower CLIP scores. However, injecting an excessive number of fine-grained anchors overwhelms the token limit of the CLIP text encoder during the diffusion process. This constraint conflict forces the denoising U-Net to produce cluttered backgrounds, sharply degrading image realism. Therefore, maintaining a moderate anchor density yields the optimal balance between strict semantic control and high-fidelity generation.

\section{Conclusion}
In this paper, we presented Semantic Anchor-aligned Multimodal Augmentation (SAMA), a unified framework designed to overcome data scarcity across heterogeneous MIE tasks. By systematically re-encoding entities, relations, and events into structured semantic anchors, SAMA effectively bridges the gap between fine-grained cross-modal alignment and task-agnostic generalization. Our approach integrates a Collaborative Multi-Experts MLLM with an Anchor-Preserving Diffusion strategy, ensuring that synthetic samples possess both linguistic diversity and strict visual identity preservation. Furthermore, the implementation of Dual-Constraint Filtering guarantees the reliability of the augmented data by rigorously enforcing semantic fidelity. Extensive experiments confirm that SAMA consistently outperforms state-of-the-art DA methods, particularly in low-resource settings, establishing a new paradigm for robust and unified MIE.

\bibliographystyle{IEEEtran}
\bibliography{ref}

\begin{IEEEbiography}[{\includegraphics[width=1in,height=1.25in,clip,keepaspectratio]{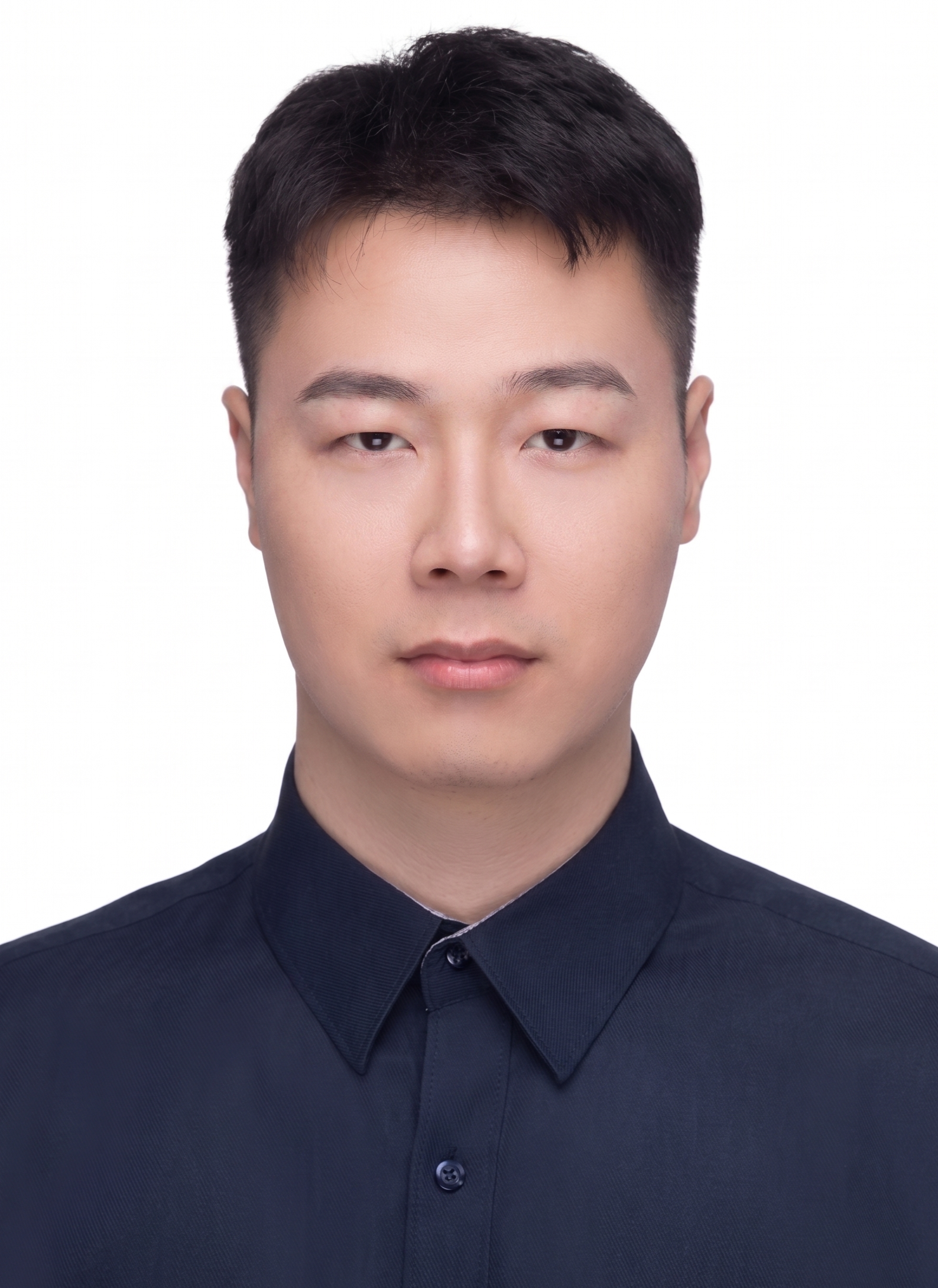}}]{Quanjiang Guo} received the BEng degree from Beijing University of Technology in 2020. He is currently a PhD candidate at University of Electronic Science and Technology of China. He has published more than 10 research articles in top-tier conferences such as AAAI, IJCAI, ACL, ICCV, ICDE, EMNLP and COLING. His current research interests include large language models, information extraction and natural language processing.
\end{IEEEbiography}

\begin{IEEEbiography}[{\includegraphics[width=1in,height=1.25in,clip,keepaspectratio]{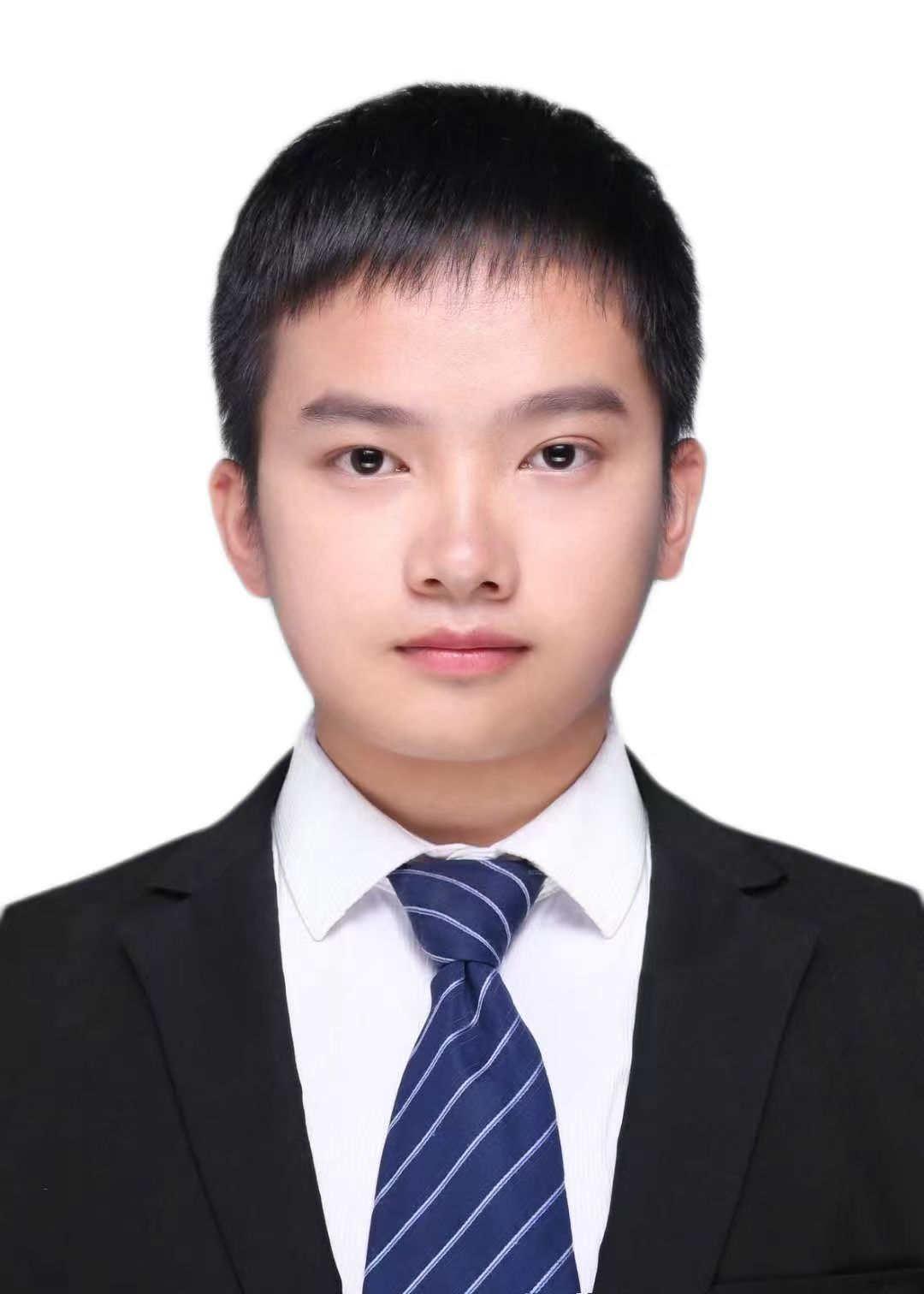}}]{Chong Mu} received the BEng degree from Hefei University of Technology (HFUT), China in 2020. He obtained his PhD degree in the School of Information and Software Engineering, University of Electronic Science and Technology of China (UESTC), China in 2024. He is currently a postdoctoral researcher at UESTC. His research interests include knowledge graph completion and inductive reasoning.
\end{IEEEbiography}
\vspace*{-0.65cm}
\begin{IEEEbiography}[{\includegraphics[width=1in,height=1.25in,clip,keepaspectratio]{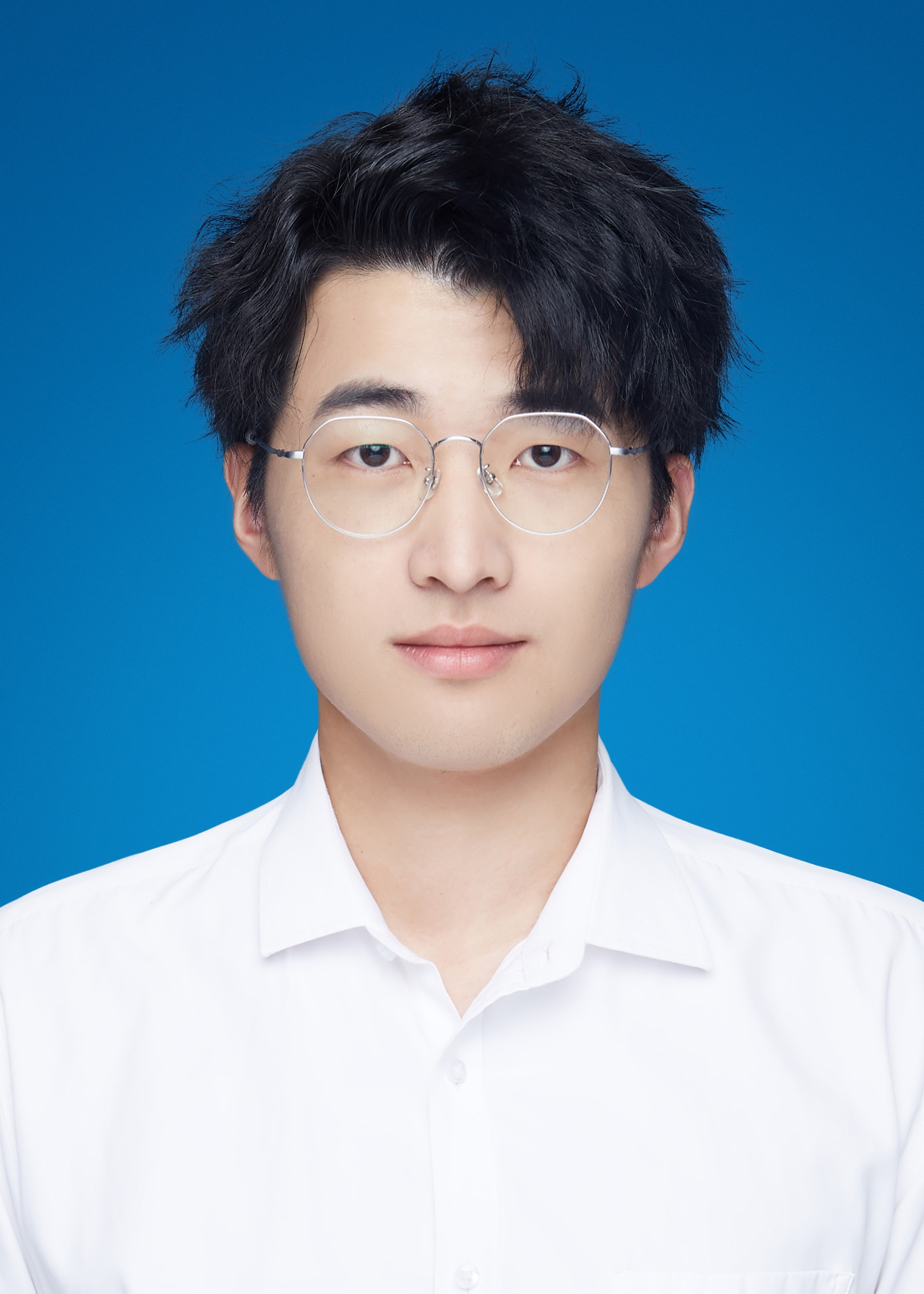}}]{Jiazhou Pan} received BSng degree in 2023 from the Jiangsu University of Science and Technology. He is currently working toward the MS degree in computer  science and engineering at the University of Electronic Science and Technology of China. His research interests include large language model, federated learning and knowledge graph.
\end{IEEEbiography}
\vspace*{-0.65cm}
\begin{IEEEbiography}[{\includegraphics[width=1in,height=1.25in,clip,keepaspectratio]{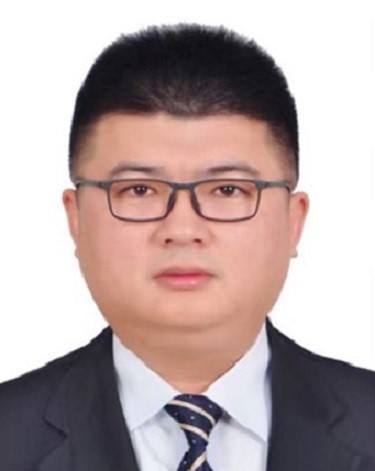}}]{Ming Jia} is currently a PhD candidate at University of Electronic Science and Technology of China. His research focuses on knowledge graph.
\end{IEEEbiography}
\vspace*{-0.65cm}
\begin{IEEEbiography}[{\includegraphics[width=1in,height=1.25in,clip,keepaspectratio]{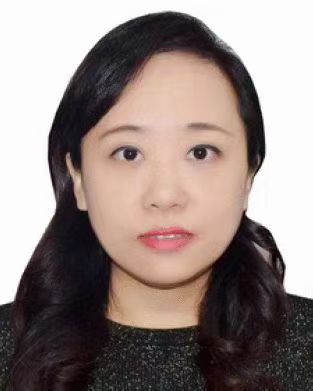}}]{Ling Tian} received the BEng, MEng, and PhD degrees in computer science from University of Electronic Science and Technology of China, China in 2003, 2006, and 2010, respectively. She is currently an full professor with the School of Computer Science and Engineering, UESTC, China. Her research interests include artificial intelligence, digital multimedia, cloud computing and big data technologies.
\end{IEEEbiography}
\vspace*{-0.65cm}
\begin{IEEEbiography}[{\includegraphics[width=1in,height=1.25in,clip,keepaspectratio]{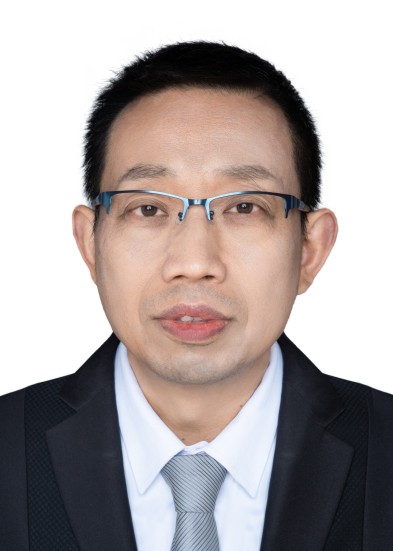}}]{Hui Gao} holds the positions of professor and doctoral supervisor at the University of Electronic Science and Technology of China. In 1991, he graduated from Peking University and obtained his Bachelor's degree. In 2005, he received his Ph.D. from the University of Groningen in the Netherlands. His research primarily focuses on machine learning and formal verification.
\end{IEEEbiography}
\vspace*{-0.65cm}
\begin{IEEEbiography}[{\includegraphics[width=1in,height=1.25in,clip,keepaspectratio]{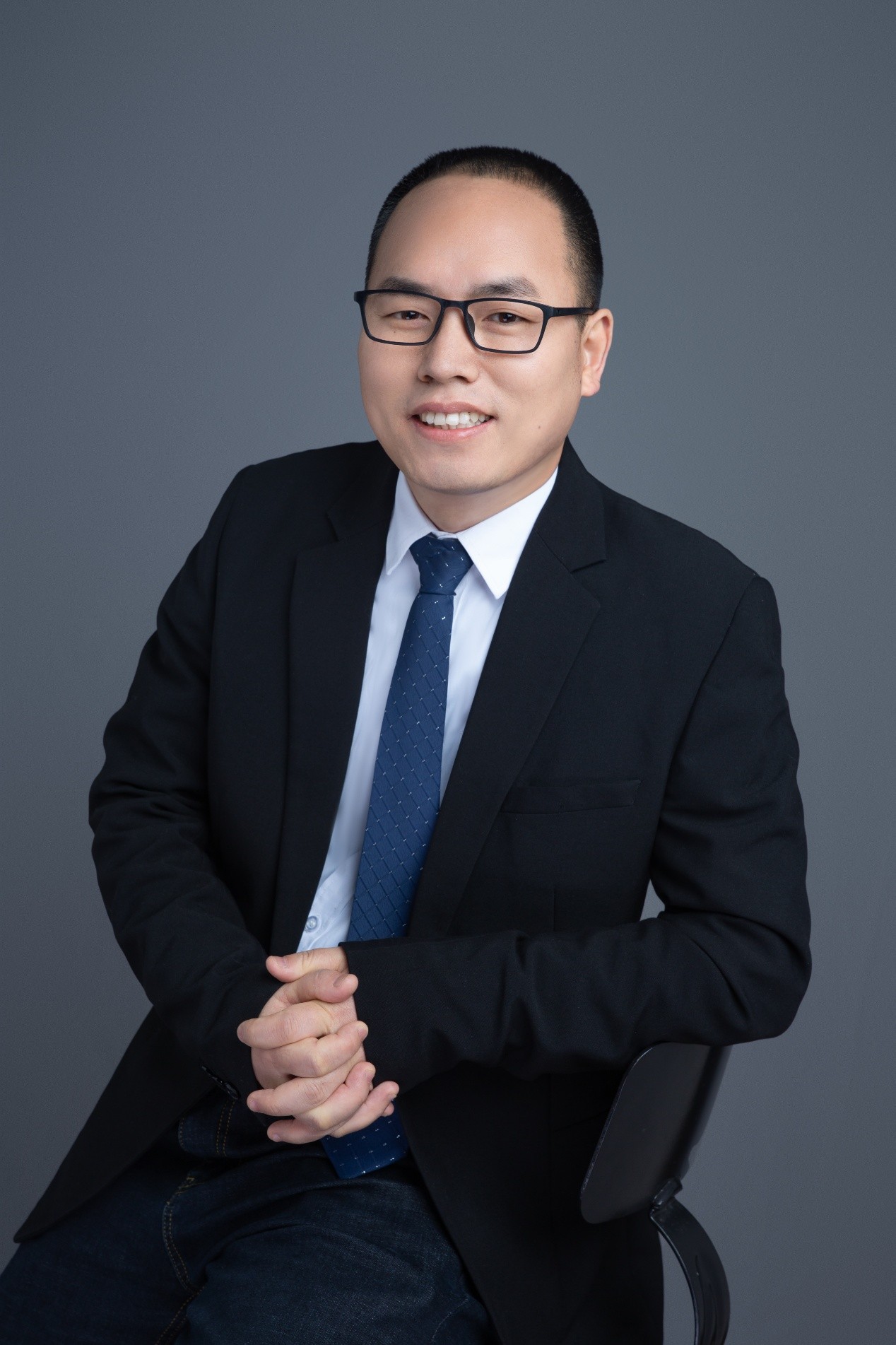}}]{Zhao Kang} received the Ph.D. degree in computer science from the Southern Illinois University Carbondale, Carbondale, IL, USA, in 2017. He is currently a Professor in the School of Computer Science and Engineering, University of Electronic Science and Technology of China, Chengdu, China. He has published more than 100 research articles in top-tier conferences and journals, including ICML, NeurIPS, ICLR, IEEE TRANSACTIONS ON CYBERNETICS, IEEE TRANSACTIONS ON IMAGE PROCESSING, IEEE TRANSACTIONS ON KNOWLEDGE AND DATA ENGINEERING, and IEEE TRANSACTIONS ON NEURAL NETWORKS AND LEARNING SYSTEMS. His research interests include graph machine learning and large language
models. He serves as an Associate Editor for Neural Networks and Pattern Recognition.
\end{IEEEbiography}

\vfill

\end{document}